\documentclass[10pt,twocolumn,letterpaper]{article}

\usepackage{cvpr}              %

\usepackage{adjustbox}
\usepackage{graphicx}
\usepackage{scalerel}
\usepackage{caption}
\usepackage{subcaption}
\usepackage{printlen}
\usepackage{float}
\usepackage{multirow}
\usepackage{wrapfig}
\usepackage{booktabs}
\usepackage{pifont}
\usepackage{makecell}
\usepackage{nicefrac}
\usepackage{epigraph}
\usepackage{url}
\usepackage{stmaryrd}
\usepackage{verbatim}
\usepackage{amsmath}
\usepackage{amssymb}
\usepackage{algorithm}
\usepackage{algpseudocode}
\usepackage{amsfonts}
\usepackage{pgf}
\usepackage{pgfplots}
\usepackage{ifthen}
\usepackage{soul}
\usepackage{tabularx}
\usepackage{tikz}
\usetikzlibrary{angles,quotes,3d,math,arrows.meta,calc,positioning,fit,backgrounds,decorations.pathreplacing,calligraphy,shapes,shapes.multipart}
\usepackage[accsupp]{axessibility}

\newcommand\rurl[1]{%
  \href{https://#1}{\nolinkurl{#1}}%
}

\newcommand{\broom}{\protect\scalerel*{\includegraphics{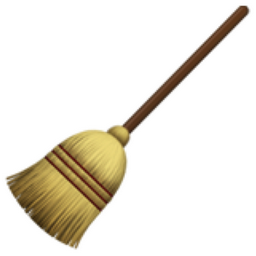}}{H}}

\newcommand{\feat}{\mathrm{feat}}
\newcommand{\feats}[2]{\feat_\mathrm{c}^{#1}(#2)}
\newcommand{\featsk}[1]{\feats{(k)}{#1}}
\newcommand{\featt}[2]{\feat^{#1}(#2)}
\newcommand{\feattk}[1]{\featt{(k)}{#1}}
\newcommand{\proj}{\mathrm{proj}}
\newcommand{\projk}{\proj^{(k)}}

\newcommand{\pck}{$\text{PCK}_{\text{bbox}}\!$@\!$\alpha$=$0.1$}

\newcommand{\x}{\mathbf{x}}
\newcommand{\eps}{\boldsymbol{\epsilon}}

\newcommand{\cmark}{\ding{51}}
\newcommand{\xmark}{\ding{55}}

\makeatletter
\newcommand{\xleftrightarrow}[2][]{\ext@arrow 3359\leftrightarrowfill@{#1}{#2}}
\newcommand{\xdashrightarrow}[2][]{\ext@arrow 0359\rightarrowfill@@{#1}{#2}}
\newcommand{\xdashleftarrow}[2][]{\ext@arrow 3095\leftarrowfill@@{#1}{#2}}
\newcommand{\xdashleftrightarrow}[2][]{\ext@arrow 3359\leftrightarrowfill@@{#1}{#2}}
\def\rightarrowfill@@{\arrowfill@@\relax\relbar\rightarrow}
\def\leftarrowfill@@{\arrowfill@@\leftarrow\relbar\relax}
\def\leftrightarrowfill@@{\arrowfill@@\leftarrow\relbar\rightarrow}
\def\arrowfill@@#1#2#3#4{%
  $\m@th\thickmuskip0mu\medmuskip\thickmuskip\thinmuskip\thickmuskip
   \relax#4#1
   \xleaders\hbox{$#4#2$}\hfill
   #3$%
}
\makeatother

\newcommand{\stefan}[1]{\textbf{{\color{ourgreen}[Stefan: #1]}}}
\newcommand{\nick}[1]{\textbf{{\color{ourturquoise}[Nick: #1]}}}

\newcommand{\suppvspreprint}[2]{#2}

\renewcommand{\nick}[1]{}
\renewcommand{\stefan}[1]{}

\definecolor{ourgreen}{RGB}{46, 204, 113}
\definecolor{ourgreenborder}{RGB}{39, 174, 96}
\definecolor{ourblue}{RGB}{52, 152, 219}
\definecolor{ourblueborder}{RGB}{41, 128, 185}
\definecolor{ourorange}{RGB}{230, 126, 34}
\definecolor{ourorangeborder}{RGB}{211, 84, 0}
\definecolor{ourred}{RGB}{231, 76, 60}
\definecolor{ourredborder}{RGB}{192, 57, 43}
\definecolor{ouryellow}{RGB}{241, 196, 15}
\definecolor{ouryellowborder}{RGB}{243, 156, 18}
\definecolor{ourpurple}{RGB}{155, 89, 182}
\definecolor{ourpurpleborder}{RGB}{142, 68, 173}
\definecolor{ourturquoise}{RGB}{26, 188, 156}
\definecolor{ourturquoiseborder}{RGB}{22, 160, 133}
\definecolor{ourturquoise}{RGB}{26, 188, 156}
\definecolor{ourturquoiseborder}{RGB}{22, 160, 133}
\definecolor{ourwhite}{RGB}{236, 240, 241}
\definecolor{ourwhiteborder}{RGB}{189, 195, 199}
\definecolor{ourgray}{RGB}{149, 165, 166}
\definecolor{ourgrayborder}{RGB}{127, 140, 141}
\definecolor{ourwhite2}{RGB}{246, 247, 248}

\definecolor{ourhighlightcolor}{RGB}{46, 204, 113}

\newcolumntype{H}{>{\setbox0=\hbox\bgroup}c<{\egroup}@{}}

\newcommand{\tikzstylenodedistance}{4mm}
\newcommand{\tikzstyleinnersep}{2mm}
\newcommand{\tikzstyleminimumheight}{8.75mm}
\newcommand{\tikzstyleminimumwidth}{12mm}

\tikzset{
    node distance=\tikzstylenodedistance,
    text centered,
    anchor=center,
}
\tikzset{
    standard node/.style n args={1}{%
        rectangle,
        rounded corners=0.1cm,
        fill=our#1,
        draw=our#1border,
        line width=0.04cm,
        minimum height=\tikzstyleminimumheight,
        minimum width=\tikzstyleminimumwidth,
        inner sep=\tikzstyleinnersep,
        text centered,
        anchor=center,
        align=center,
    }
}
\tikzset{
    standard node module/.style n args={0}{%
        rectangle,
        rounded corners=0.1cm,
        fill=ourturquoise,
        draw=ourturquoiseborder,
        line width=0.04cm,
        minimum height=\tikzstyleminimumheight, %
        minimum width=12mm, %
        inner xsep=\tikzstyleinnersep,
        inner ysep=1mm,
        text centered,
        anchor=center,
        align=center,
    }
}
\tikzset{
    standard node image/.style n args={1}{%
        rectangle,
        fill=our#1,
        draw=our#1border,
        line width=0.04cm,
        minimum height=\tikzstyleminimumheight,
        minimum width=\tikzstyleminimumwidth,
        inner sep=0,
        text centered,
        anchor=center,
        align=center,
    }
}
\tikzset{
    standard node circle/.style n args={1}{%
        fill=our#1,
        draw=our#1border,
        circle,
        inner sep=0.1cm,
        minimum height=0,
        minimum width=0,
    }
}
\tikzset{
    standard node circle/.prefix style = standard node
}

\tikzset{
    standard line/.style n args={0}{%
        line width=0.04cm,
        rounded corners=0.1cm,
    }
}
\tikzset{
    standard arrow/.style n args={0}{%
        -latex,
    }
}
\tikzset{
    standard arrow/.prefix style = standard line
}

\tikzset{
    simple node image/.style n args={0}{%
        rectangle,
        inner sep=0,
        text centered,
        anchor=center,
        align=center,
        node distance=0mm
    }
}

\definecolor{cvprblue}{rgb}{0.21,0.49,0.74}
\usepackage[pagebackref,breaklinks,colorlinks,allcolors=cvprblue]{hyperref}
\usepackage[capitalize,noabbrev]{cleveref}

\def\paperID{} %
\def\confName{CVPR}
\def\confYear{2025}

\title{\broom{} CleanDIFT: Diffusion Features without Noise}

\author{
    Nick Stracke\thanks{} \qquad Stefan Andreas Baumann\footnotemark[1] \qquad Kolja Bauer\footnotemark[1] \qquad Frank Fundel \qquad Bj\"orn Ommer\\
    CompVis @ LMU Munich, MCML\\
    {\tt\small \{nick.stracke,b.ommer\}@lmu.de}
}

\begin{document}

\twocolumn[{%
    \maketitle
    \vspace{-11mm}
    \begin{center}
        \rurl{compvis.github.io/cleandift}
        \vspace{2mm}
        \captionsetup{type=figure}{\includegraphics[width=0.8\textwidth]{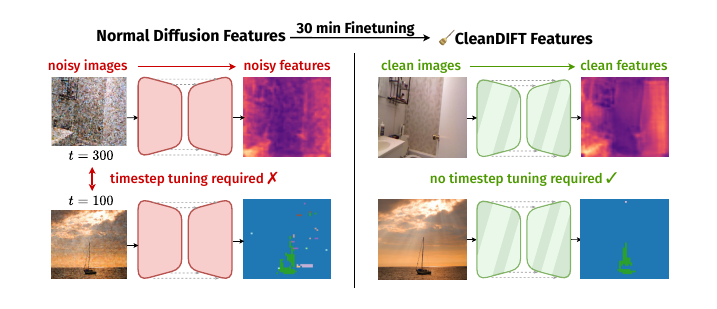}}
        \vspace{-6mm}
        \captionof{figure}{Our proposed CleanDIFT feature extraction method yields noise-free, timestep-independent, general-purpose features that significantly outperform standard diffusion features. CleanDIFT operates on clean images, while extracting diffusion features with existing approaches requires adding noise to an image before passing it through the model. Adding noise reduces the information present in the image and requires tuning a timestep per downstream task.}
        \label{fig:teaser}
    \end{center}
    \vspace{0mm}
}]

\def\thefootnote{*}\footnotetext{Equal Contribution}
\begin{abstract}
Internal features from large-scale pre-trained diffusion models have recently been established as powerful semantic descriptors for a wide range of downstream tasks. Works that use these features generally need to add noise to images before passing them through the model to obtain the semantic features, as the models do not offer the most useful features when given images with little to no noise. We show that this noise has a critical impact on the usefulness of these features that cannot be remedied by ensembling with different random noises. 
We address this issue by introducing a lightweight, unsupervised fine-tuning method that enables diffusion backbones to provide high-quality, noise-free semantic features. We show that these features readily outperform previous diffusion features by a wide margin in a wide variety of extraction setups and downstream tasks, offering better performance than even ensemble-based methods at a fraction of the cost.

\end{abstract}

\section{Introduction}
\label{sec:intro}

Learning meaningful visual representations that capture a vast amount of world knowledge remains a key problem in the field of computer vision. Diffusion models can be trained at scale in a self-supervised manner and have rapidly advanced the state of the art in image~\cite{dhariwal2021diffusion, nichol2021glide, rombach2022high} and video generation~\cite{ho2022video, singer2023make, esser2023structure}, making them a good candidate to learn visual representations. Many early works have already achieved impressive results using internal features from large-scale pretrained diffusion models for a wide variety of tasks, such as semantic correspondence detection~\cite{tang_emergent_2023, zhang_tale_2023, zhang_telling_2024}, semantic segmentation~\cite{baranchuk2021label, wu2023diffumask, namekata2023emerdiff}, panoptic segmentation~\cite{xu2023open}, object detection~\cite{chen2023diffusiondet}, and classification~\cite{li2023your}. However,  the optimal approach to extract this world knowledge from a diffusion model remains uncertain.

\begin{figure}[t]
    \centering
    \includegraphics[width=\columnwidth]{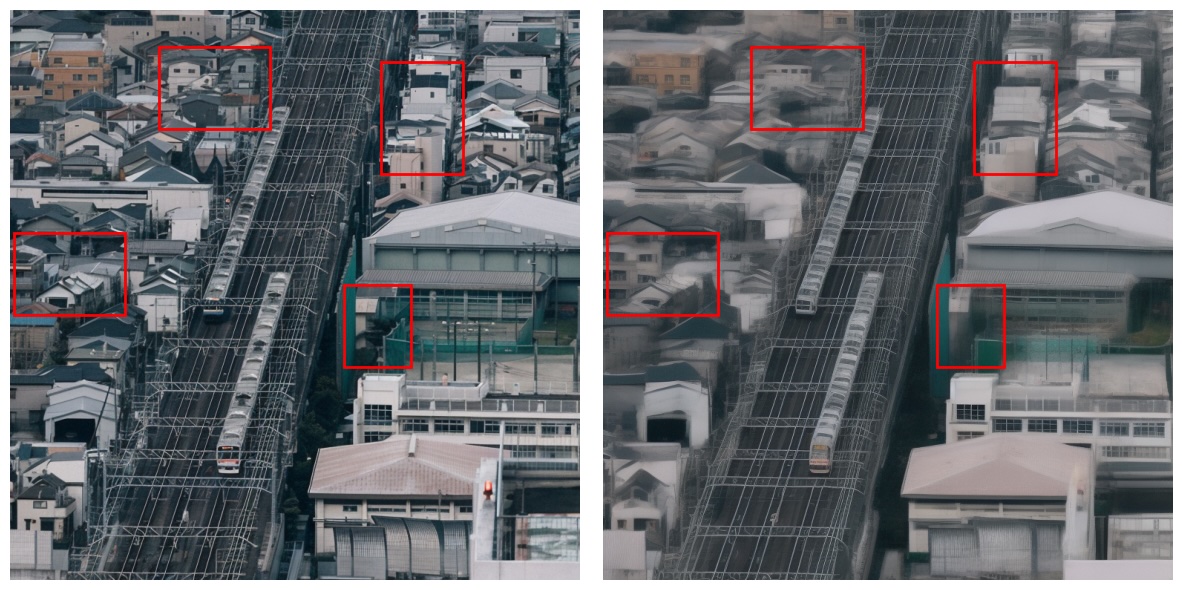}
    \adjustbox{max width=\columnwidth}{
        \begin{minipage}{1.4\columnwidth}
            \begin{tabularx}{\linewidth}{XX}
                 \makecell[c]{(a) Reconstruction without Noise} & \makecell[c]{(b) Diffusion Model Reconstruction\\with added noise ($t = 261$ \cite{tang_emergent_2023})}
            \end{tabularx}
        \end{minipage}
    }
    \caption{\textbf{Deterioration of Diffusion Features}. As current methods \textit{need} to pass noisy images to the model to obtain useful features, they significantly reduce the information available. We alleviate this problem by obtaining useful features without noise, improving the performance of downstream tasks.}
    \label{fig:information-bottleneck}
\end{figure}

To understand why that is the case, we take a look at how diffusion models are trained: a varying amount of noise is added to a clean input image (forward process) and the model is tasked to remove the noise from the image (backward process). The amount of added noise is dependent on the diffusion \textit{timestep}. 
As a result, the model learns to operate on noisy images and also becomes \textit{dependent} on the noise timestep as different noise levels require the model to perform different tasks \cite{biroli2024dynamical,balaji2022ediff}. Since noisy images inherently contain less information than clean images (cf.~\cref{fig:information-bottleneck}), we hypothesize that this harms the internal feature representation of diffusion models~\cite{chen2024deconstructing} and, thus, the extractable world knowledge. Furthermore, the timestep acts as a hyperparameter that influences the internal feature representation and needs to be picked independently for every downstream application (cf.~\cref{fig:teaser-fig-appendix}). 

We propose a novel feature extraction method that (1) eliminates the need to destroy information by adding noise to the input; and (2) produces timestep-independent generic diffusion features useful for a wide range of down-stream tasks, alleviating the need to tune a noising timestep per down-stream task.
We show how to adapt an off-the-shelf large-scale pre-trained diffusion backbone to provide these features at minimal cost (approximately 30 minutes of fine-tuning on a single A100 GPU) and demonstrate improved performance across a wide range of downstream tasks.

We achieve this by viewing a diffusion model as a family of $T$ feature extractors that operate on images with different noise levels and provide features with different characteristics. 
We consolidate all $T$ feature extraction functions in our feature extractor by aligning their internal representations.
Specifically, we initialize our feature extractor as a trainable copy of the diffusion model; fine-tune it with clean images and no timestep input; and align its features with all $T$ time-dependent feature extractors of the diffusion model.

We evaluate our improved features across a wide variety of downstream tasks, such as semantic correspondence matching, monocular depth estimation, semantic segmentation, and classification, and find that they consistently improve upon approaches based on standard diffusion features. These improvements are most evident for dense visual tasks such as semantic correspondence matching, where our features show substantial performance gains across a wide variety of setups~\cite{tang_emergent_2023,zhang_telling_2024,zhang_tale_2023} and set a new state-of-the-art for unsupervised semantic correspondence matching. Additionally, our proposed method eliminates the need for noise or timestep ensembling~\cite{tang_emergent_2023}, offering substantial speed gains (e.g., $8\times$ over DIFT~\cite{tang_emergent_2023}), on top of improved quality. Our method is generic and integrates easily with established methods, such as fusing diffusion and DINOv2 features~\cite{zhang_tale_2023, zhang_telling_2024}.

\noindent Our main contributions are as follows:
\begin{enumerate}
    \item We propose CleanDIFT, a finetuning approach for diffusion models that enables them to operate on clean images and makes the inherent world knowledge of these models more accessible.
    \item We show how to consolidate information from all diffusion timesteps into a single feature prediction, removing the need for task-specific timestep tuning.   
    \item We demonstrate significant performance gains of our diffusion feature extraction technique across a wide range of down-stream tasks, notably surpassing the current state of the art in zero-shot unsupervised semantic correspondence detection. We further demonstrate the generality of our enhanced features by showing that these performance gains transfer to advanced methods that fuse diffusion features or operate in a supervised setting.
    \item Our proposed approach is significantly more efficient than previous methods that tried to address this problem by noise ensembling or supervised training.
\end{enumerate}

\section{Related Work}

\paragraph{Self-Supervised Representation Learning}

Features from large, pre-trained foundation models have been shown to yield competitive performance to supervised models for a variety of downstream tasks, both in zero-shot and fine-tuning settings \cite{oquab2024dinov2, radford2021learning, he2022masked}. These foundation models are trained on different pre-text tasks like inpainting \cite{he2022masked}, predicting transformations \cite{gidaris2018unsupervised}, patch reordering \cite{noroozi2016unsupervised, misra2020self}, and discriminative tasks \cite{caron2021emerging, oquab2024dinov2}. DINOv2~\cite{oquab2024dinov2} uses a discriminative objective combined with self-distillation to learn general-purpose visual features that have proven useful for a variety of downstream tasks~\cite{darcet2024vision}. CLIP \cite{radford2021learning} learns such features by employing a contrastive objective on text-image pairs. Masked Autoencoders~\cite{he2022masked} (MAEs) are trained to reconstruct masked out patches of the input, also resulting in general-purpose visual features.

\paragraph{Diffusion Models as Self-Supervised Learners}

Diffusion models~\cite{song2019generative, ho2020denoising, song2020score} are generative models that have defined the state-of-the-art in image generation~\cite{dhariwal2021diffusion, nichol2021glide, rombach2022high, saharia2022photorealistic, podellsdxl, esser2024scaling}, video generation~\cite{esser2023structure, polyak2024movie}, and audio generation~\cite{kong2021diffwave, evansfast} in recent years. Their primary purpose is to generate high-quality samples (images, videos, etc.). However, generation can also be interpreted as a pretext task for learning expressive features, since the model has to build up comprehensive world knowledge in order to generate plausible samples \cite{tang_emergent_2023,fuest2024diffusion_rl_survey,chen2024deconstructing,li2023dreamteacher, hudson2024soda}. Features from diffusion models (typically referred to as \textit{diffusion features}) are obtained by passing a noised image through the diffusion model and extracting intermediate feature representations. They have been shown to be useful for a variety of tasks such as finding semantic correspondences~\cite{tang_emergent_2023, hedlin2024unsupervised, luo_diffusion_2024}, semantic and panoptic segmentation~\cite{baranchuk2021label, wu2023diffumask, xu2023open, namekata2023emerdiff},~classification~\cite{li2023your}, and object detection~\cite{chen2023diffusiondet}. 

For semantic correspondence matching, features are leveraged to identify semantically matching regions across images. Existing approaches utilize diffusion features either in a zero-shot setting~\cite{tang_emergent_2023, hedlin2024unsupervised} or fine-tune them for the semantic correspondence task~\cite{luo_diffusion_2024, li2024sd4match, zhang_telling_2024}. Some zero-shot approaches do not fine-tune on semantic correspondence but still require tuning a prompt to activate attention maps at the query location of the correspondence~\cite{hedlin2024unsupervised}. In contrast, our approach aims to provide universal features usable for various down-stream tasks in a true zero-shot manner. 

Further, diffusion features have been shown to complement features from other self-supervised learning methods such as DINOv2~\cite{zhang_tale_2023,oquab2024dinov2, fundel2025distilldift}. DINOv2 features provide sparse but accurate semantic information, while Diffusion features provide dense spatial information albeit with sometimes inaccurate semantics. State-of-the-art approaches for semantic correspondence detection exploit this and fuse features~\cite{zhang_tale_2023, zhang_telling_2024}. Compared to DINOv2 features, diffusion features yield smoother, spatially more coherent correspondences~\cite{zhang_tale_2023}.

\vspace{-2mm}

\paragraph{Distillation for Diffusion Models} 

Knowledge Distillation~\cite{hinton2015distilling} is a technique used to distill knowledge from a teacher model into a student model. In the context of diffusion models, this is typically applied either to reduce the required denoising steps~\cite{salimans2022progressive} or to distill a classifier-free guided~\cite{ho2022classifier} model into one without CFG~\cite{meng2023distillation}.
While our approach is inspired by distillation, we consolidate features from $T$ different models, with $T$ being the number of discrete diffusion timesteps, because the features are different at every timestep.

\section{Method}

\subsection{Preliminaries}

\paragraph{Diffusion Models}

Diffusion models are trained to predict a clean image $\x_0$ given a noisy image $\x_t$, either explicitly or implicitly. The noisy image $\x_t$ is a weighted sum with random Gaussian noise $\x_t = \sqrt{\alpha_t}\x_0 + \sqrt{1-\alpha_t} \eps$ with timestep-dependent coefficients $\alpha_t$ and noise $\eps \sim \mathcal{N}(0, \textbf{I})$. $t \in [0, T]$ denotes the time step of the diffusion process, with $t=0$ corresponding to the clean image and $t=T$ corresponding to pure noise.

Intuitively, the model faces different objectives for different noise levels \cite{biroli2024dynamical,balaji2022ediff}: for very high noise, there is little information in the input, and the model is first generating the coarse structure of the image \cite{rissanen2022generative}. At lower noise levels, more high- and medium-frequency information is available and the task shifts to generating finer details and intricate structures.
This multi-objective nature intuitively explains why previous methods found diffusion features extracted from different timesteps to provide information with differing semantics.

\paragraph{Diffusion Feature Extraction}
Typically, diffusion feature extraction happens after first adding noise to an image and passing the resulting 
$\x_t$ to a U-Net~\cite{ronneberger2015unet} denoiser. Features are then extracted at multiple hand-picked locations of the U-Net decoder \cite{tang_emergent_2023, zhang_tale_2023, zhang_telling_2024, baranchuk2021label, namekata2023emerdiff}. Different levels of noise added to the input image result in features beneficial for different downstream applications. Typical diffusion timesteps are $t = 261$~\cite{tang_emergent_2023}, $t = 100$~\cite{zhang_tale_2023} or $t=50$~\cite{zhang_telling_2024}.
By adding noise to the input image, these methods bottleneck the perceptual information the model can extract. To illustrate this, we show Stable Diffusion 2.1's reconstruction of an image noised at $t = 261$~\cite{tang_emergent_2023} in \cref{fig:information-bottleneck}.

\begin{figure}[t]
    \centering
    \scalebox{.6}{\includegraphics[]{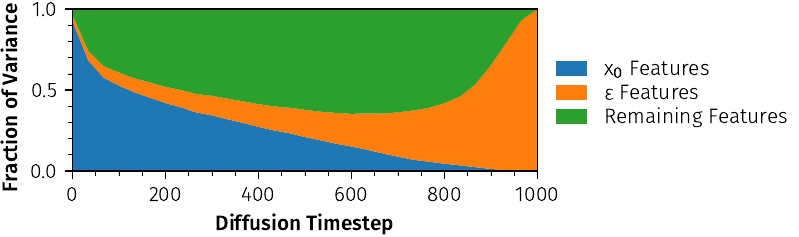}}
    \caption{Fraction of variance of diffusion features explained by 1) encoding the clean image at $t=0$ (no additive noise), and 2) encoding just the added noise $\eps$ at $t = 999$. Even at relatively low timesteps such as $t = 261$ as used by DIFT~\cite{tang_emergent_2023}, a substantial part of the features directly depends only on the added noise.}
    \label{fig:diffusion_features_variance_explained}
\end{figure}

\paragraph{Diffusion Features Encode Noise} We hypothesize that diffusion features extracted from noisy images $\x_t$ encode the noise $\eps$ in addition to information from the image. We investigate this hypothesis in a very simple setting by examining how well features $\feat(\eps; T)$ extracted from only the noise $\eps$ approximate the features $\feat(\x_t = \sqrt{\alpha_t}\x_0 + \sqrt{1-\alpha_t} \eps; t)$. Using least squares, we fit a single scalar approximation coefficient to obtain the optimal reconstruction. We then quantify how much of the variance of the overall features is explained by this approximation (cf.~\cref{fig:diffusion_features_variance_explained}). Even at relatively low timesteps, such as $t = 261$ used by DIFT~\cite{tang_emergent_2023}, encoding pure noise explains a substantial fraction of the features' variance. Current diffusion feature methods extract this information jointly with the image information. Our proposed method addresses this issue by eliminating the noise from the feature extraction process.

We further analyze the residual and similarly decompose it via the features predicted for the clean image $\feat(x_t = x_0; t=0)$. We find that they do not fully explain the remainder of the features either. Instead, a substantial part of the feature variance at medium noise timesteps is timestep dependent and cannot be attributed to components present at $t = 0$ or $t = T$. This matches observations by previous works~\cite{tang_emergent_2023} that found diffusion features at higher timesteps offer better semantics, despite the added noise at the image input.

\subsection{\broom{}CleanDIFT: Noise-Free Diffusion Features}

\begin{figure}[t]
    \centering
    \includegraphics[width=\columnwidth]{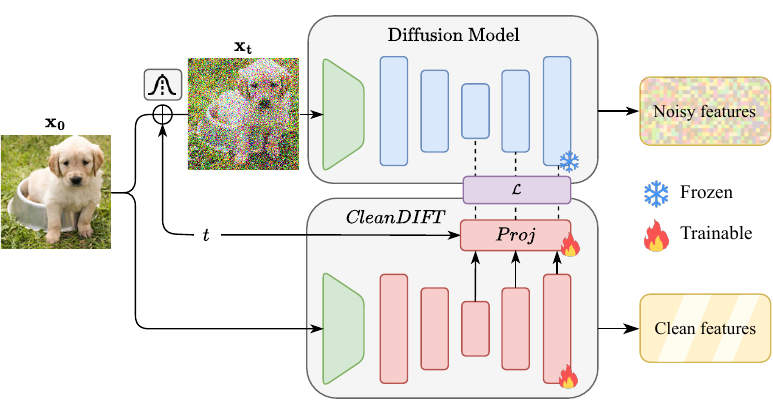}
    \caption{Our training setup. We train our model to predict features from a clean input image, while the frozen diffusion model is fed the noisy image. The projection heads project our model's features onto the noisy diffusion model features, given the noising timestep $t$. For downstream tasks, we discard the projection heads and directly use our model's internal representations as features.\vspace{-4mm}}
    \label{fig:model-architecture}
\end{figure}

We present CleanDIFT, our method to address the problem of noisy and time-dependent diffusion features. \textit{CleanDIFT} extracts \underline{clean} \underline{DI}ffusion \underline{F}ea\underline{T}ures from a pretrained diffusion backbone through a lightweight fine-tuning process. An overview of our setup is shown in \cref{fig:model-architecture}. 

\paragraph{Extraction Setup}
We train our feature extraction model to match the diffusion model's internal representations.
We initialize the feature extraction model as a trainable copy of the diffusion model. Crucially, the feature extraction model is given the clean input image, while the diffusion model receives the noisy image and the corresponding timestep as input. 
Our goal is to obtain a single, noise-free feature map from the feature extraction model that consolidates the information of the diffusion model's timestep-dependent internal representations into a single one. To align our model's representations with the timestep-dependent diffusion model features during training, we introduce point-wise timestep-conditioned feature projection heads. The feature maps predicted by these projection heads are then aligned to the diffusion model's features.
For feature extraction at inference time, we usually discard the projection heads and directly use the feature extraction model's internal representations. However, the projection heads can also be used to efficiently obtain feature maps for specific timesteps by reusing the feature extraction model's internal representations and passing them through the projection heads for different $t$ values.

\paragraph{Training Objective}
We regard the diffusion model as a family of feature extraction functions $\mathrm{feat}(\cdot, \eps, t)$ for timestep $t\in[1, 999]$ and noise $\eps \sim \mathcal{N}(0, \mathbf{I})$. Each of these functions maps an image $\x$ to a feature vector $\mathrm{feat}(\x, \eps, t)$.
We aim to consolidate the information provided by all feature extraction functions into a single joint function $\mathrm{feat}_c(\cdot)$ with the same dimensionality:
{
    \setlength{\parskip}{0pt}
    \setlength{\textfloatsep}{0.5pt}
    \setlength{\floatsep}{1pt}
    \setlength{\intextsep}{1pt}
    \begin{figure}[H]
        \centering
        \begin{tikzpicture}
            \node[] (a) {$\left.\begin{array}{c}
                \mathrm{feat}(\x, \eps, t=\ \ \ \ 1)\\
                \mathrm{feat}(\x, \eps, t=\ \ \ \ 2)\\
                \vdots\\
                \mathrm{feat}(\x, \eps, t=999)
            \end{array}\right\}$};
            \node[right=14mm of a] (b) {$\feats{}{\x}$};
            \draw[-stealth,line width=0.25mm] ($(a.east) + (-2mm, 0)$) --node[above] {\footnotesize consolidate} node[below] {\footnotesize information} (b);
            \node[above=0mm of a,xshift=-2mm] (teacher) {\footnotesize \underline{Stable Diffusion}};
            \node[above=0mm of a,xshift=38mm] (student) {\footnotesize \broom{} \underline{CleanDIFT}};
        \end{tikzpicture}
    \end{figure}
}
\noindent To align our model's features with the diffusion model's features, we maximize the similarity between the diffusion model's features and the projected features of our feature extraction model:
{
    \setlength{\parskip}{0pt}
    \setlength{\textfloatsep}{0.5pt}
    \setlength{\floatsep}{1pt}
    \setlength{\intextsep}{1pt}
    \begin{figure}[H]
        \centering
        \adjustbox{max width=\columnwidth}{
        \begin{tikzpicture}
            \node[] (a) {$\feats{}{\x}$};
            \node[right=0mm of a] (b) {$
            \left\{\begin{array}{c}
                \proj(\ \cdot\ , t=\ \ \ \ 1) \xrightarrow{\ \ \ }\mathcal{L}\xleftarrow{\ \ \ } \mathrm{feat}(\x, \eps, t=\ \ \ \ 1)\\
                \proj(\ \cdot\ , t=\ \ \ \ 2) \xrightarrow{\ \ \ }\mathcal{L}\xleftarrow{\ \ \ } \mathrm{feat}(\x, \eps, t=\ \ \ \ 2)\\
                
                \vdots\\
                \proj(\ \cdot\ , t=999) \xrightarrow{\ \ \ }\mathcal{L}\xleftarrow{\ \ \ } \mathrm{feat}(\x, \eps, t=999)
            \end{array}\right.
            $};
            \draw[-stealth,line width=0.25mm] ($(a.east) + (0mm, 0)$) -- ($(b.west) + (2mm, 0)$);
        \end{tikzpicture}
        }
    \end{figure}
}
\noindent Specifically, we minimize the negative cosine similarity between the diffusion model's features and our model's features extracted at stages $k = \{1, ..., K\}$ in the network. Given a clean image $\x_0$, the feature extraction model's output for feature map $k$ is denoted as $\featsk{\x_0}$. Our CleanDIFT feature map is then adapted by the learned projection heads $\projk(\featsk{\x_0}, t)$, where $\projk{}(\cdot, \cdot)$ is the projection head for feature map $k$. The diffusion model receives the noisy image $\x_t$ corresponding to the same $\x_0$ and timestep $t$. The projection head then learns a timestep-dependent alignment from CleanDIFT features to the diffusion model's features $\feattk{\x_t; t}$. Putting it all together, our loss function is defined as:
\begin{equation}
    \mathcal{L} = - \!\!\!\sum_{k=1}^{K} \mathrm{sim}(\projk(\featsk{\x_{0}}; t), \feattk{\x_{t}; t}).
\end{equation}
For each training image $\x_0$, we sample $I$ different noising timesteps $t_i$ in a stratified manner, with each timestep
$t_i\sim\mathcal{U}(\frac{i}{I}T, \frac{i+1}{I}T)$,
where $T$ is the maximum timestep. By sampling multiple timesteps per image we incentivize the feature extraction model to match the diffusion model's features across the entire noise spectrum.

\section{Experiments}
\label{sec:exps}
We test our hypothesis that the proposed extraction setup enables us to leverage more of the world knowledge inherent in diffusion models compared to existing diffusion feature extraction methods while being task-agnostic and timestep-independent. To that end, we evaluate our features on a wide range of downstream tasks: unsupervised zero-shot semantic correspondence, monocular depth estimation, semantic segmentation, and classification. We compare our features against standard diffusion features, methods that combine diffusion features with additional features, and non-diffusion-based approaches.

\subsection{Experimental Setup}

\paragraph{Implementation Details} 
Following previous works~\cite{tang_emergent_2023, zhang_tale_2023, zhang_telling_2024}, we evaluate our method on a Stable Diffusion (SD) backbone~\cite{rombach2022high}. We apply our method to SD 1.5 and SD 2.1 to enable fair comparisons with existing methods that use either. We fully fine-tune our feature extraction model on image-caption pairs for only 400 steps, taking 30 minutes on a single A100 GPU, which was sufficient for our strong performance and further training did not yield any significant gains. We extract features after the U-Net's middle block and after each of the U-Net's decoder blocks, except the two final blocks. A detailed visualization of where we extract features is provided in~\cref{fig:architecture-details}. This yields a total of $K = 11$ feature maps that we align between the diffusion model and the feature extraction model. Our point-wise feature projection heads consist of three stacked Feed Forward Networks (FFNs) that are zero-initialized such that initially they act as identity mappings due to their residual connections. Since every aligned feature map has its own projection head, this results in 45M additional trainable parameters for SD 2.1.  We study the effect of different projection head architectures in~\cref{sec:proj-heads}.
We train using Adam~\cite{kingma2015adam} with a batch size of 8 and a learning rate of 2e-6 with a linear warmup. For stratified timestep sampling, we utilize $I=3$ stratification bins across all our experiments, i.e. three different noise levels per training image. 

\paragraph{Datasets}

We fine-tune our feature extraction model on a random subset of COYO-700M~\cite{kakaobrain2022coyo-700m}, which is similar to the LAION~\cite{schuhmann2022laion}, the dataset that Stable Diffusion 1.5 and 2.1 were trained on originally. That way, we ensure that all performance improvements originate from the feature extraction model consolidating the diffusion model's internal feature representations over time, not from choosing a different dataset that matches the test dataset distribution more closely.
The subset selects images with a minimum size of $512^2$. We crop and resize them to match the corresponding input resolution of the underlying diffusion model. We analyze the effect of using different datasets in \cref{supp:dataset}.

\subsection{Unsupervised Semantic Correspondence}

\begin{figure}[t]
    \centering
    \includegraphics[width=\columnwidth]{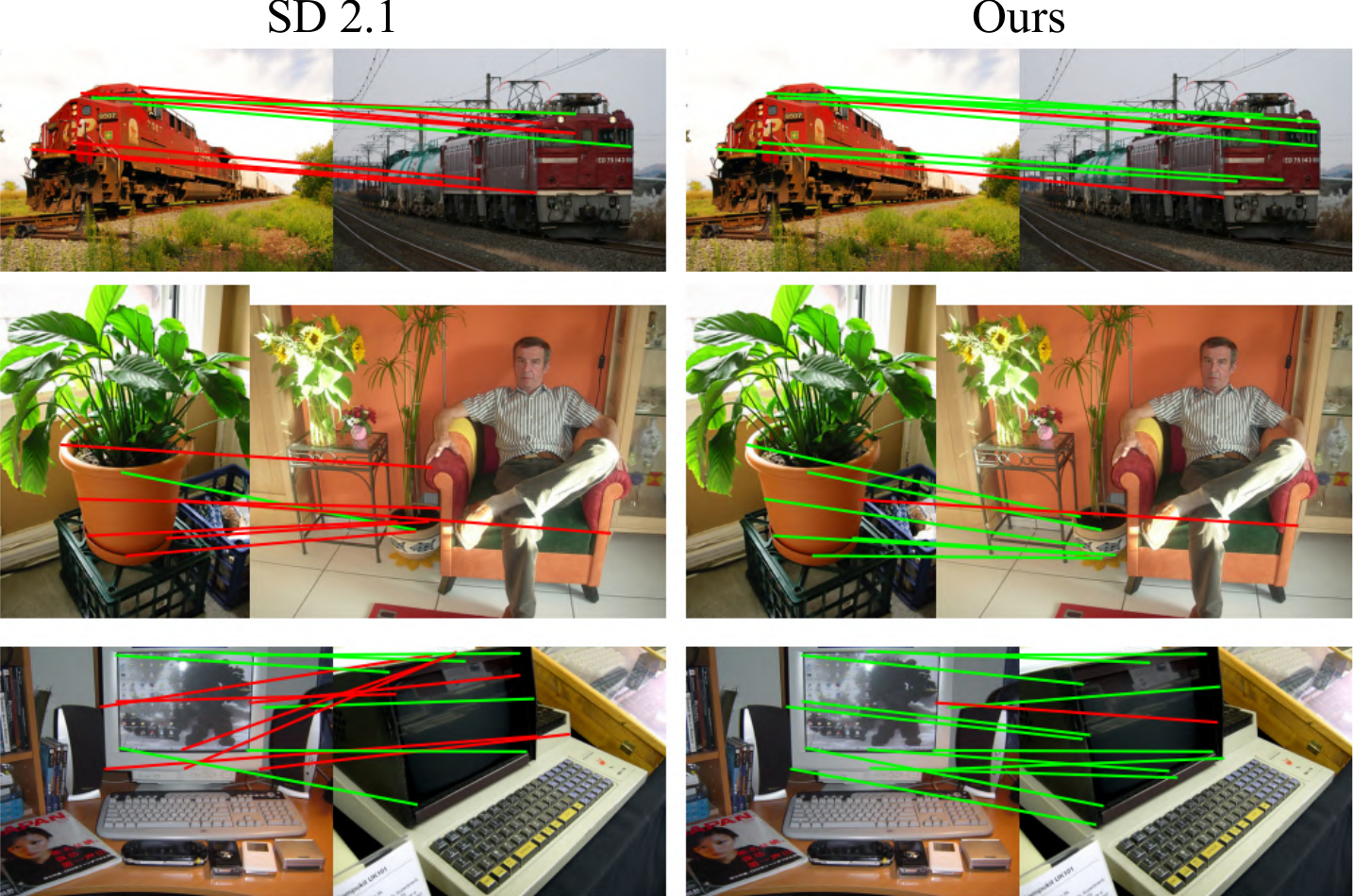}
    \vspace{-4mm}
    \caption{Semantic correspondence results using DIFT~\cite{tang_emergent_2023} features with the standard SD 2.1 ($t=261$) and our CleanDIFT features. Our clean features show significantly less incorrect matches than the base diffusion model. \vspace{-4mm}}
    \label{fig:correspondence-examples}
\end{figure}

As many previous diffusion feature methods focus on (unsupervised) semantic correspondence matching~\cite{tang_emergent_2023,zhang_tale_2023,zhang_telling_2024,luo_diffusion_2024}, we perform an extensive evaluation of our method on this task.
Following previous works on semantic correspondence matching~\cite{tang_emergent_2023, zhang_tale_2023, zhang_telling_2024}, we measure our performance in Percentage of Correct Keypoints (PCK). We average PCK directly across all keypoints, not over images. We use $\alpha=0.1$ as a threshold and report both PCK values with error margins relative to the image size and to the bounding box size, denoted as $\text{PCK}_{\text{img}}$ and $\text{PCK}_{\text{bbox}}$ respectively. We evaluate the performance on the test split of the SPair-71k dataset, which consists of approximately 12k image pairs from 18 categories. Some existing works~\cite{tang_emergent_2023, zhang_tale_2023, zhang_telling_2024} evaluate on additional datasets but find SPair-71k to be the most challenging and therefore the most informative benchmark. For the text prompt we use ``A photo of a \texttt{category}.'', with \texttt{category} being the corresponding category of the SPair image. We experiment with distilling the text conditioning in~\cref{sec:add-quant-eval}.

\begin{figure}[t]
    \centering
    \scalebox{.6}{\includegraphics[]{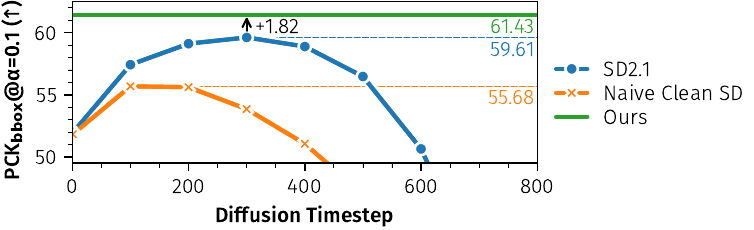}}
    \caption{Following~\cite{tang_emergent_2023}, we evaluate semantic correspondence matching accuracy for different noise levels. Our feature extractor outperforms the standard noisy diffusion features across all timesteps $t$. We additionally demonstrate that simply providing the diffusion model with a clean image and a non-zero timestep does not result in improved performance. \vspace{-5mm}}
    \label{fig:correspondences-over-timesteps}
\end{figure}

\vspace{-2mm}
\paragraph{Results} We first compare our extracted features to DIFT~\cite{tang_emergent_2023}, an approach that detects semantic correspondences using standard diffusion features. Substituting these with our CleanDIFT features yields a performance increase of $1.79$ absolute percentage points for $\text{PCK}_{\text{img}}$ and $1.86$ percentage points for $\text{PCK}_{\text{bbox}}$. Notably, DIFT averages the extracted feature maps across 8 different noise samples. Without this averaging over noise samples, our performance gain is even larger (2.81 PCK@$\alpha_\textrm{img}$ gain). This indicates that our feature extraction model learns more than a mere averaging over the noise in the diffusion model's feature maps (see \cref{fig:correspondence-examples} and \cref{sec:supp-add-qual} for examples). We present an extended version of the time-step dependent performance analysis conducted by~\cite{tang_emergent_2023} in~\cref{fig:correspondences-over-timesteps}: We evaluate the diffusion model's performance for different timesteps $t$ in two settings. In the first setting, we provide the diffusion model with a noisy input image $\x_t$ as usual. In the second setting, we demonstrate that feeding the clean image along with a non-zero timestep is not a viable solution to obtain meaningful features: We provide the diffusion model with the clean input image $\x_0$ for all timesteps $t$. We observe that the model's performance for the clean input image degrades faster and has a lower peak than for the noisy input. This is to be expected, as the diffusion model was trained on noisy images, not clean images. 
Importantly, our CleanDIFT features are timestep-independent and consistently outperform standard diffusion features, even when the latter are optimized for the best-performing timestep. We further observe that this advantage generalizes effectively to other backbones, such as DiTs~\cite{peebles2023scalable} (see~\cref{tab:supp:more-backbones}).

A Tale of Two Features~\cite{zhang_tale_2023} extends the approach of DIFT by combining diffusion features with DINOv2~\cite{oquab2024dinov2} features. Again, we replace the standard diffusion features with our CleanDIFT features and observe that the performance gain transfers when combining our features with DINOv2 features.
Telling Left from Right~\cite{zhang_telling_2024} further improves upon the results of A Tale of Two Features by introducing a test-time adaptive pose alignment strategy. We observe that the performance gain transfers to this setting as well. To the best of our knowledge, Telling Left from Right combined with our CleanDIFT features sets a new state-of-the-art in unsupervised zero-shot semantic correspondence matching. In summary, replacing standard diffusion features with our CleanDIFT features consistently results in a significant performance improvement across all three methods. We show an overview of the results in~\cref{tab:correspondence-quantitative} and a more extensive evaluation per category in~\cref{sec:add-quant-eval}.

\begin{table}[h]
\centering

\adjustbox{max width=\linewidth}{

\begin{tabular}{lcHll}
\toprule
\multirow{2}{*}[-3pt]{Method} & \multirow{2}{*}[-3pt]{\shortstack{Our\\Features}} & \multirow{2}{*}[-3pt]{Ensemble ($\downarrow$)} & \multicolumn{2}{c}{PCK@$\alpha\ (\uparrow)$ } \\ \cmidrule(lr){4-5}
                        &&& $\alpha_\mathrm{img}$ = 0.1 & $\alpha_\mathrm{bbox}$ = 0.1   \\ \midrule
\multicolumn{4}{l}{\textit{General Approaches}}\\
DINOv2+NN                & - & ? & -          & 55.6               \\
\midrule
\multicolumn{4}{l}{\textit{Diff. Feat.-based Approaches}}\\
\multirow{2}{*}{DIFT \cite{tang_emergent_2023}}
 & {\color{ourred}\xmark} & 8$\times$  & 66.53         & 59.57               \\
& \cellcolor{ourwhite}{\color{ourgreen}\cmark} & \cellcolor{ourwhite}\textbf{1$\times$} & \cellcolor{ourwhite}68.32\textsubscript{\color{ourgreen}$\blacktriangle$1.79}  & \cellcolor{ourwhite}61.43\textsubscript{\color{ourgreen}$\blacktriangle$1.86}         \\
\multirow{2}{*}{A Tale of Two Features \cite{zhang_tale_2023}} & {\color{ourred}\xmark} & ? & 72.31          & 63.73              \\
& \cellcolor{ourwhite}\textbf{\color{ourgreen}\cmark} & \cellcolor{ourwhite}\textbf{1$\times$} & \cellcolor{ourwhite}73.35\textsubscript{\color{ourgreen}$\blacktriangle$1.04}  & \cellcolor{ourwhite}64.81\textsubscript{\color{ourgreen}$\blacktriangle$1.08}         \\
\multirow{2}{*}{Telling Left from Right \cite{zhang_telling_2024}} & {\color{ourred}\xmark} & ? & \underline{77.07}         & \underline{68.64}               \\
& \cellcolor{ourwhite}{\color{ourgreen}\cmark} & \cellcolor{ourwhite}\textbf{1$\times$} & \cellcolor{ourwhite}\textbf{78.40}\textsubscript{\color{ourgreen}$\blacktriangle$1.33}           & \cellcolor{ourwhite}\textbf{69.99}\textsubscript{\color{ourgreen}$\blacktriangle$1.35}           \\ 

\bottomrule
\end{tabular}
}
\caption{Zero-shot unsupervised semantic correspondence matching performance comparison on SPair71k \cite{min2019spair}. Our improved features consistently lead to substantial improvements in matching performance. We report PCK on the test split of SPair71k, aggregated per point. Numbers are reproduced, for a discussion and comparison to reported numbers view~\cref{tab:reported_vs_reproduced}.\vspace{-4mm}}
\label{tab:correspondence-quantitative}
\end{table}

We also investigate the performance of our features in a supervised fine-tuning setting for semantic correspondence matching. Following \cite{luo_diffusion_2024}, we train an aggregation network that uses all extracted feature maps and learns to aggregate them into a single task-specific feature map for semantic correspondence matching. In contrast to~\cite{luo_diffusion_2024}, we do not have to perform costly DDIM inversion \cite{song2021denoising} to obtain a matching noisy image for every timestep. Instead, we directly feed the clean image to our feature extraction model. Therefore, extracting features with our CleanDIFT approach is 50x faster, since we perform a single denoiser forward pass while~\cite{luo_diffusion_2024} perform 50 for the inversion. Our model achieves a $\text{PCK}_{\text{img}}$ value of $72.48$ vs their $72.75$ and a $\text{PCK}_{\text{bbox}}$ value of $64.37$ vs their $64.53$. We observe a slight performance regression compared to their approach, however, at a speedup of 50$\times$. Luo et al.~\cite{luo_diffusion_2024} also present a single-step ablation of their full method that only requires a single forward pass which makes it more comparable to ours. We outperform this single-step version by a wide margin of $9.0$ percentage points for $\text{PCK}_{\text{img}}$ and $9.1$ percentage points for $\text{PCK}_{\text{bbox}}$.

\subsection{Depth Estimation}
\label{subsec:depth}

We also investigate monocular depth estimation on NYUv2~\cite{silberman2012nyuv2}. Similar to \cite{oquab2024dinov2}, we follow the evaluation protocol from \cite{li2024binsformer}. We use SD 2.1 as the base model and resize the input to the model's native resolution of $768^2$. We extract features from the same location as \cite{tang_emergent_2023} and obtain a feature map of dimension $48^2$. Unlike \cite{oquab2024dinov2}, we do not upsample the features and directly apply the linear probe. The probe predicts depth in 256 uniform bins which we combine with a classification loss after a linear normalization following \cite{bhat2021adabins}. We train one probe for our CleanDIFT features and one for standard diffusion features at $t = 299$, as that timestep minimizes the error in our settings. Our qualitative results (see \cref{fig:depth_estimation_qualitative}) show a substantial fidelity gap in the estimated depth maps between the features from the standard SD 2.1 backbone and the features from our feature extraction model. This is reflected in a substantial improvement in quantitative metrics over the baseline as seen in \cref{tab:depth_quantitative}. Lastly, we reuse the probe trained on standard diffusion features and apply it on the CleanDIFT features. While this does not match the performance of the CleanDIFT probe, it still achieves significantly better results when compared to using standard diffusion features. This indicates that our features can be used as a drop-in replacement for the original diffusion features and offer improved performance on downstream applications.

{
\setlength{\tabcolsep}{0.01\linewidth}
\begin{figure}[t]
    \centering
    \vspace{-1.2mm}
    \begin{tabular}{c c c c}
        \footnotesize Input & \footnotesize Ours & \footnotesize $\text{SD 2.1}_{t=299}$  & \footnotesize Ground Truth \\
        \includegraphics[width=0.225\linewidth]{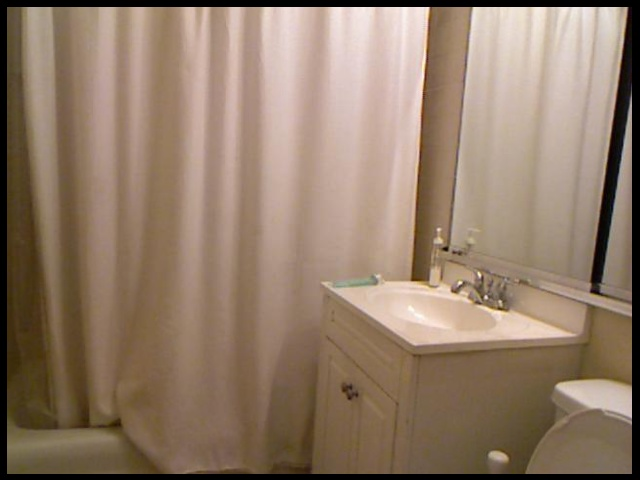} &
        \includegraphics[width=0.225\linewidth]{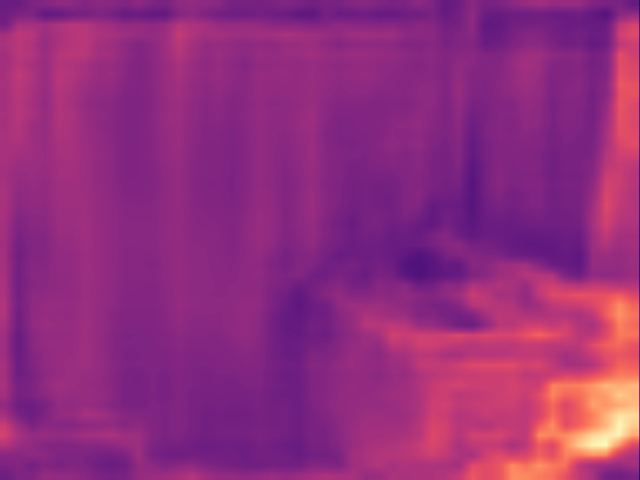}&
        \includegraphics[width=0.225\linewidth]{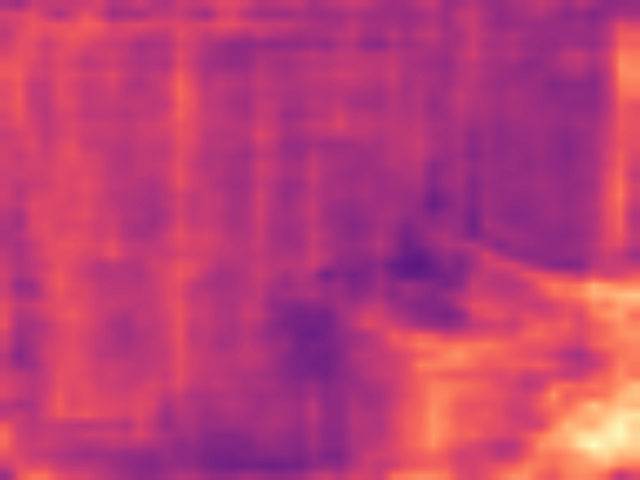} & 
        \includegraphics[width=0.225\linewidth]{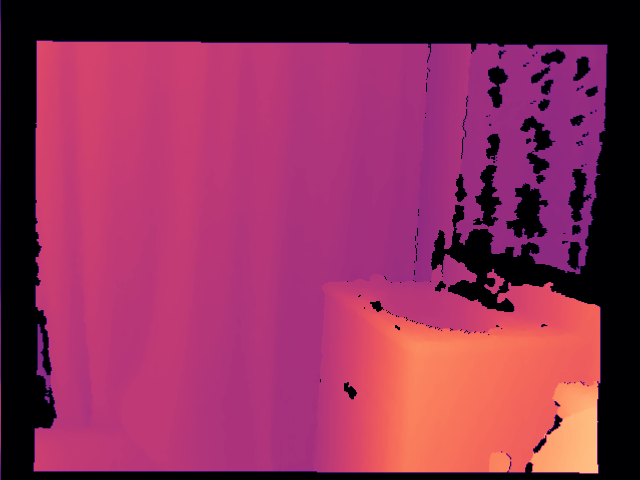} \\

        \includegraphics[width=0.225\linewidth]{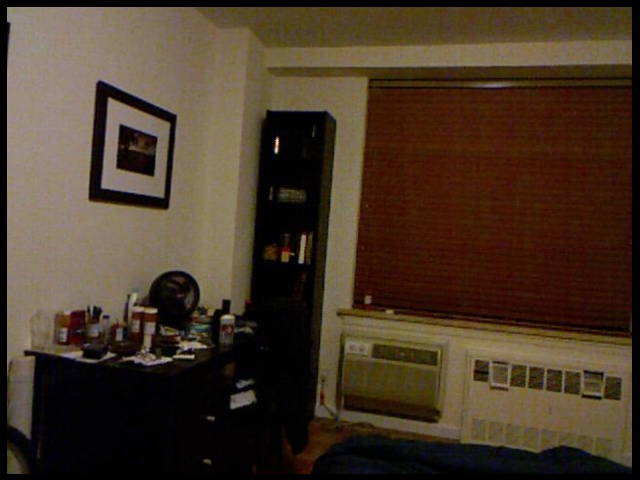} &
        \includegraphics[width=0.225\linewidth]{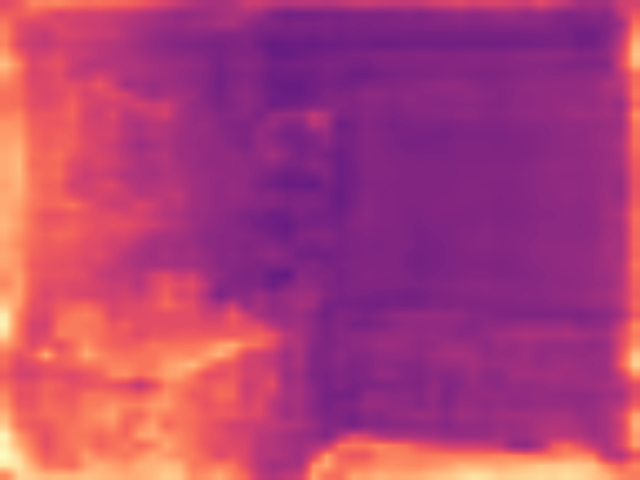}&
        \includegraphics[width=0.225\linewidth]{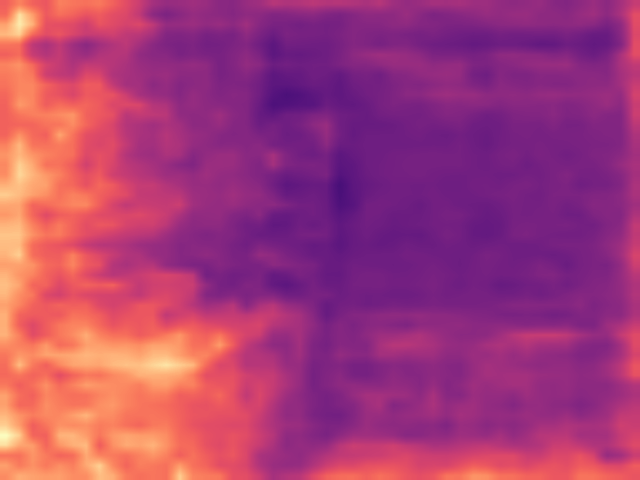} & 
        \includegraphics[width=0.225\linewidth]{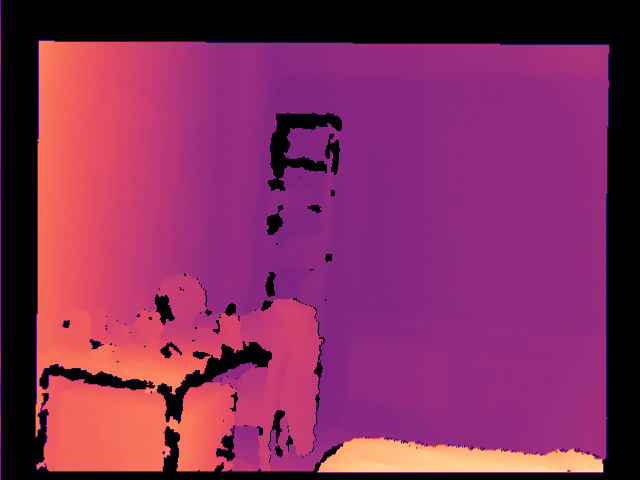} \\

        \includegraphics[width=0.225\linewidth]{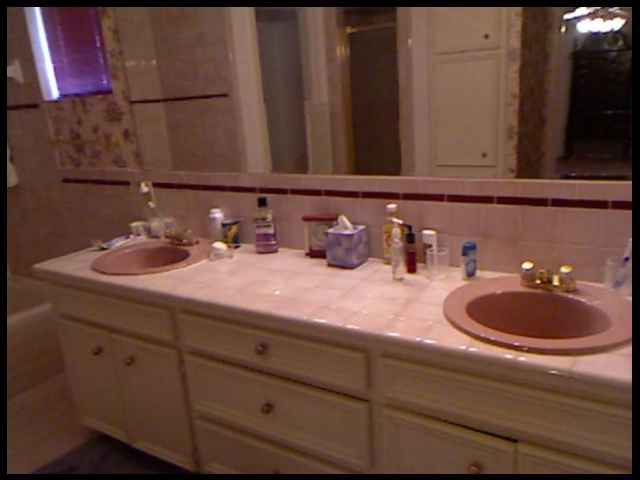} &
        \includegraphics[width=0.225\linewidth]{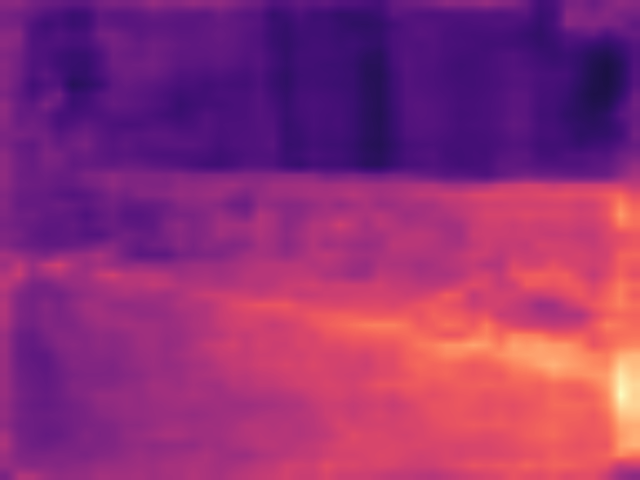}&
        \includegraphics[width=0.225\linewidth]{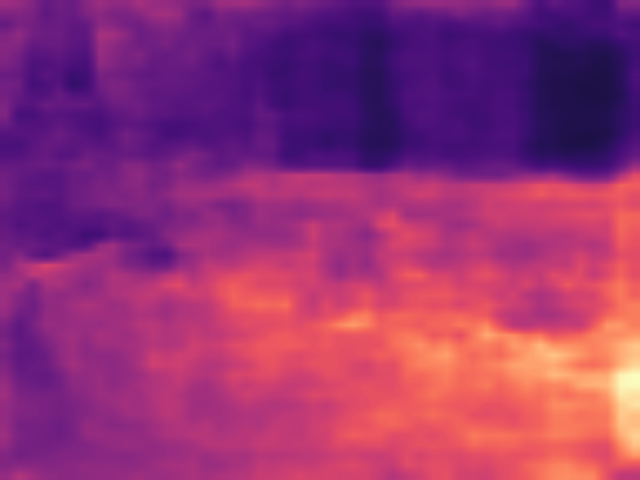} & 
        \includegraphics[width=0.225\linewidth]{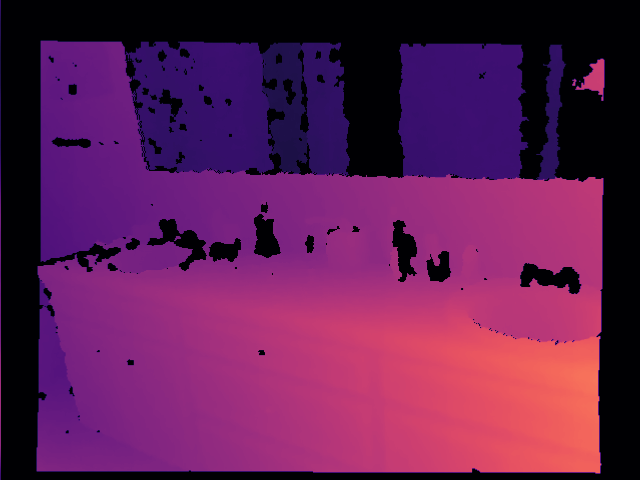} \\
    \end{tabular}
    \caption{Qualitative results for depth estimation using a linear probe on diffusion features on NYUv2 \cite{silberman2012nyuv2}. Our CleanDIFT features enable substantially better depth estimation than standard diffusion features. Note how the CleanDIFT features are far less noisy when compared to the standard diffusion features.}
    \label{fig:depth_estimation_qualitative}
\end{figure}
}

\begin{table}[t]
\centering

\adjustbox{max width=\linewidth}{
\begin{tabular}{lll}
\toprule
Method & Backbone & {RMSE ($\downarrow$)} \\
\midrule
\multicolumn{3}{l}{\textit{Self-Supervised Methods}}\\
OpenCLIP~\cite{ilharco_gabriel_2021_5143773} & ViT-G/14 & 0.541 \\
MAE~\cite{he2022masked} & ViT-H/14 & 0.517 \\
DINO~\cite{caron2021emerging} & ViT-B/8 & 0.555 \\
iBOT~\cite{zhou2022image} & ViT-L/16 & \underline{0.417} \\
DINOv2~\cite{oquab2024dinov2} & ViT-g/14 & \textbf{0.344} \\
\midrule
\multicolumn{3}{l}{\textit{Diffusion Features}}\\
\multirow{3}{*}{DIFT-like~\cite{tang_emergent_2023}} & SD 2.1~\cite{rombach2022high} & {0.469} \\
& \cellcolor{ourwhite}\textbf{Ours} & \cellcolor{ourwhite}{$\textbf{0.444}_{\color{ourgreen}\blacktriangledown0.025}$} \\
& {\ + Probes from noisy features} & $\underline{0.453}_{\color{ourgreen}\blacktriangledown0.016}$ \\
\bottomrule
\end{tabular}
}
\caption{Monocular Depth Estimation. Following \cite{oquab2024dinov2}, we evaluate metric depth prediction on NYUv2~\cite{silberman2012nyuv2} using a linear probe. Our clean features outperform the noisy features by a significant margin. Probes trained on the noisy features can be reused for the clean features, but incur a smaller performance gain.\vspace{-4mm}}
\label{tab:depth_quantitative}
\end{table}

\subsection{Semantic Segmentation}
\label{subsec:sem_seg}

To further investigate the difference between standard noisy diffusion features and our CleanDIFT features, we evaluate on the semantic segmentation task by training linear probes on our CleanDIFT features and on standard diffusion features. We utilize SD2.1 as the diffusion backbone and extract features at the same location as \cite{tang_emergent_2023}. This procedure yields feature maps of size $48^2$. We train our linear probe on the $48^2$ feature maps and upscale the obtained segmentation masks using nearest neighbor upsampling. We train and evaluate on the PASCAL VOC dataset~\cite{Everingham10}. Following common practice~\cite{namekata2023emerdiff}, we use mean Intersection over Union (mIOU) as the evaluation metric. Qualitative results are shown in \cref{fig:sem_seg}. Using our features, we observe significantly less noisy segmentations than with standard diffusion features. We show a quantitative comparison of our CleanDIFT feature's performance against standard diffusion features across timesteps in~\cref{fig:sem_seg_quantitative}. Notably, the optimal timestep appears to be around $t\!=\!100$, in contrast to the optimal timestep for semantic correspondences, which~\cite{tang_emergent_2023} found to be $t=261$. This highlights the need for tuning a timestep individually per downstream task. Our method both alleviates the need for such a timestep tuning and outperforms the standard diffusion features for the optimal timestep.

{
\setlength{\tabcolsep}{0.005\linewidth}
\begin{figure}[t]
    \centering
    \begin{minipage}{0.48\linewidth}
        \centering
        \renewcommand{\arraystretch}{0.7}
        \begin{tabular}{c c c}
            \footnotesize Input & \footnotesize Ours & \footnotesize SD 2.1 \\
            \includegraphics[width=0.32\linewidth]{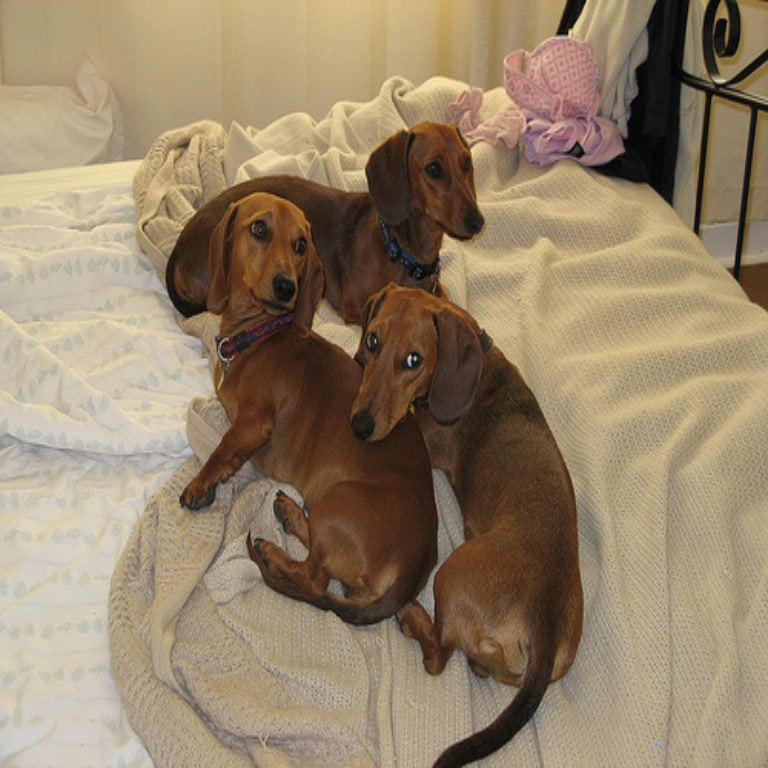} &
            \includegraphics[width=0.32\linewidth]{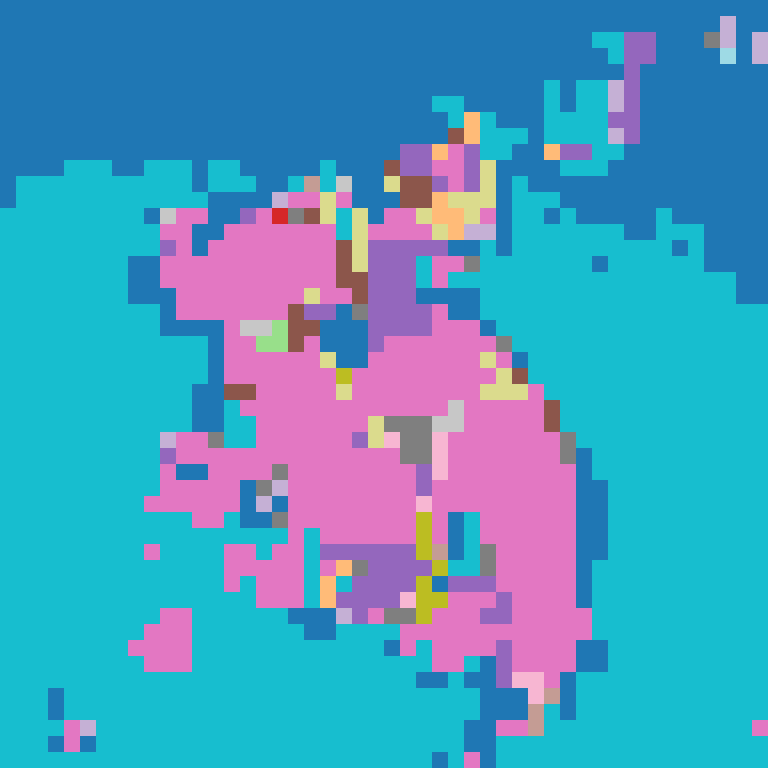} &
            \includegraphics[width=0.32\linewidth]{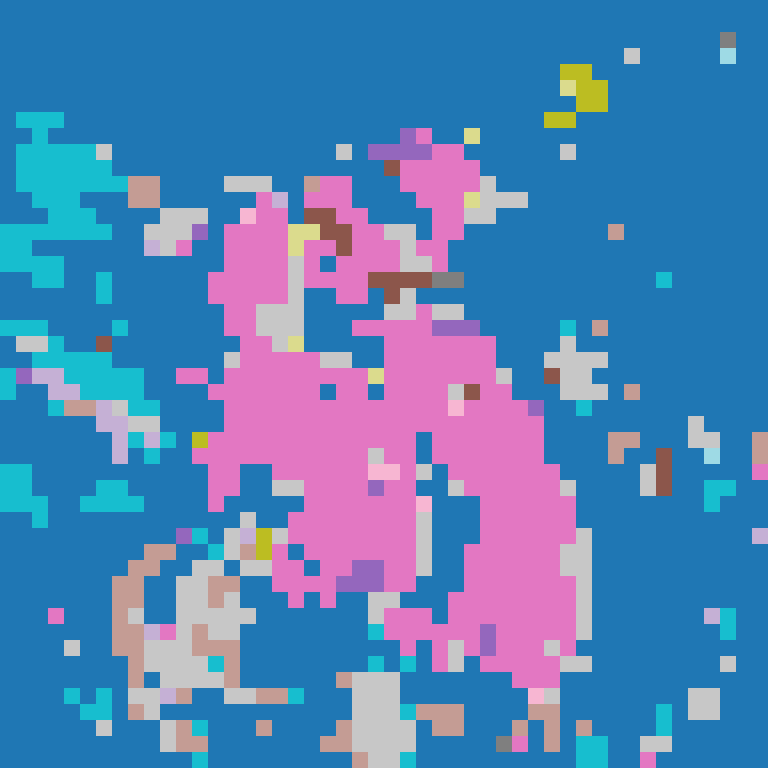} \\
            \includegraphics[width=0.32\linewidth]{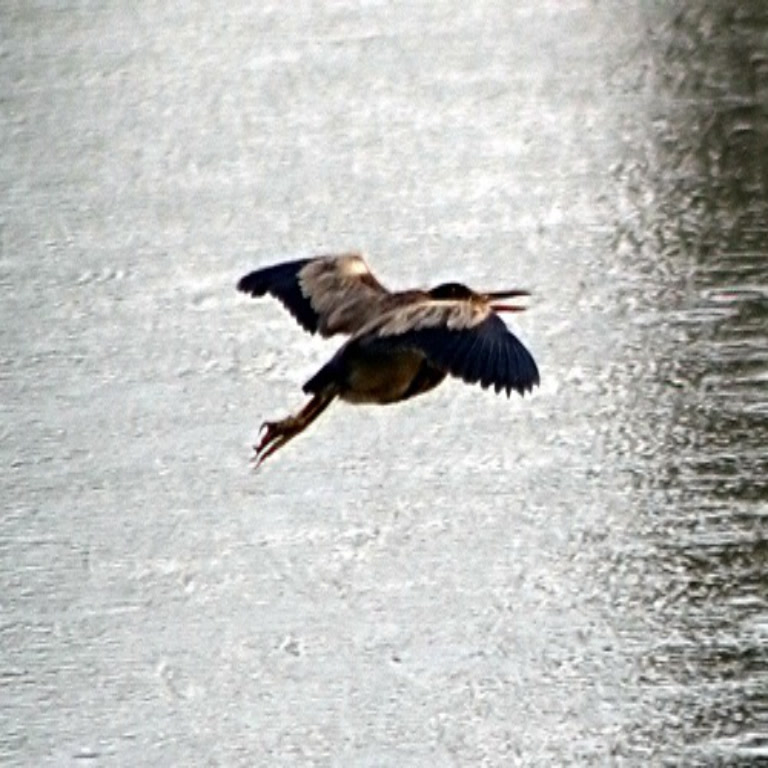} &
            \includegraphics[width=0.32\linewidth]{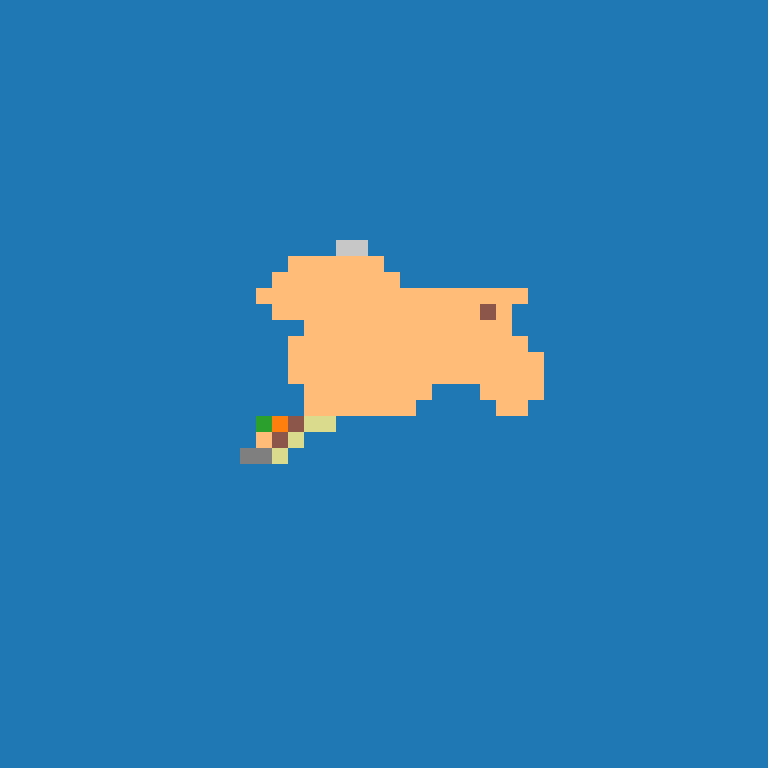} &
            \includegraphics[width=0.32\linewidth]{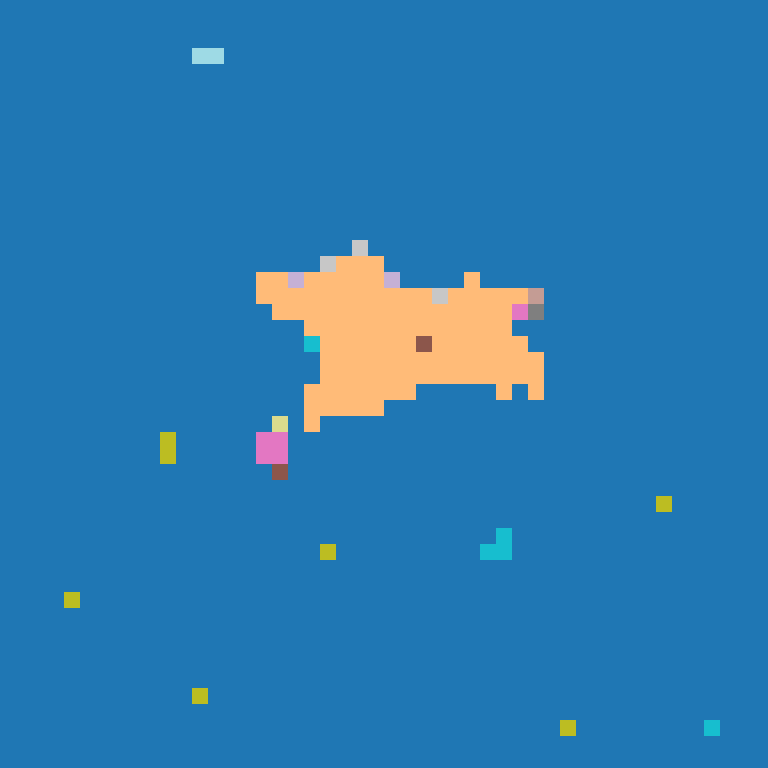} \\
            \includegraphics[width=0.32\linewidth]{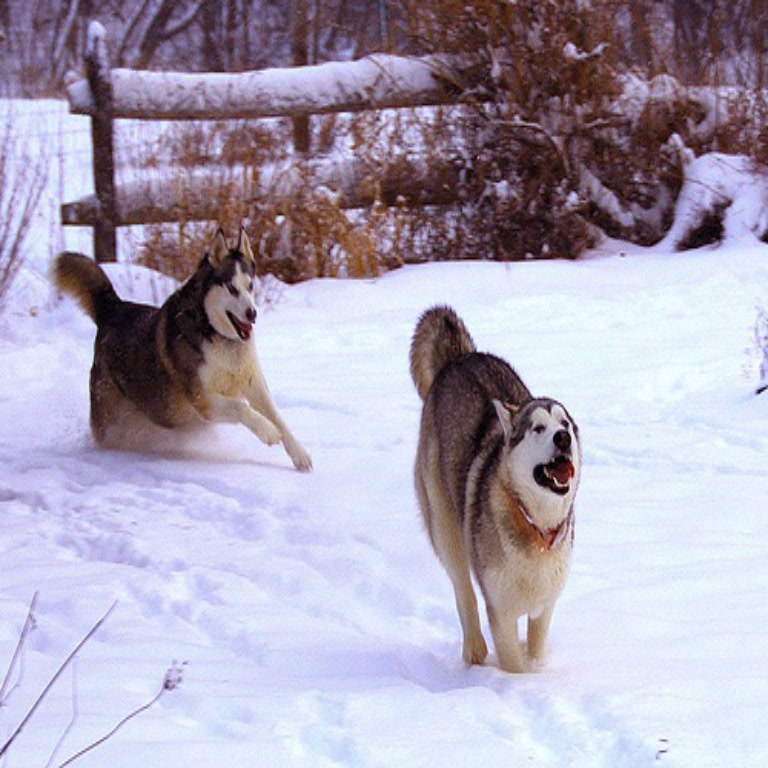} &
            \includegraphics[width=0.32\linewidth]{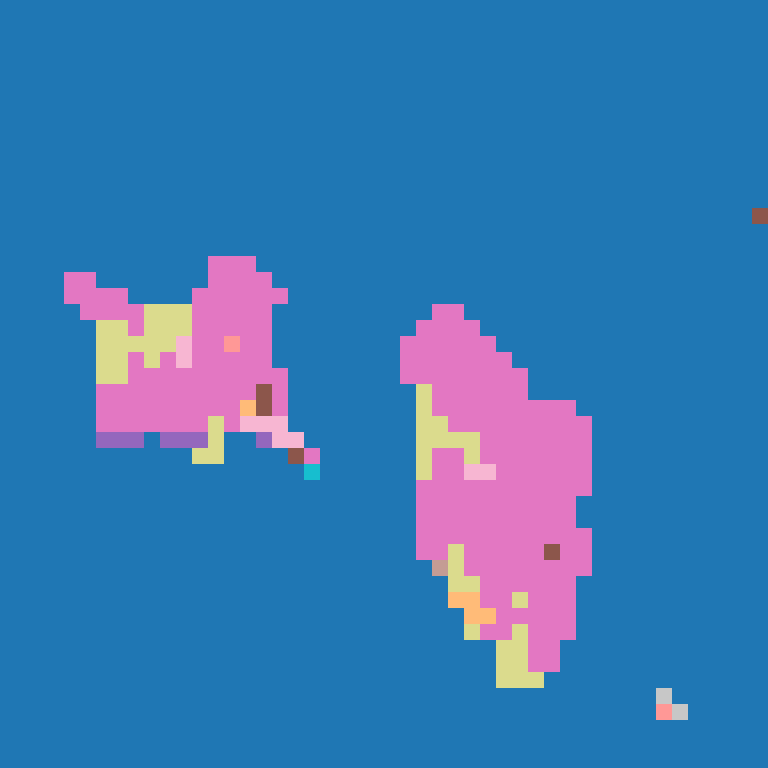} &
            \includegraphics[width=0.32\linewidth]{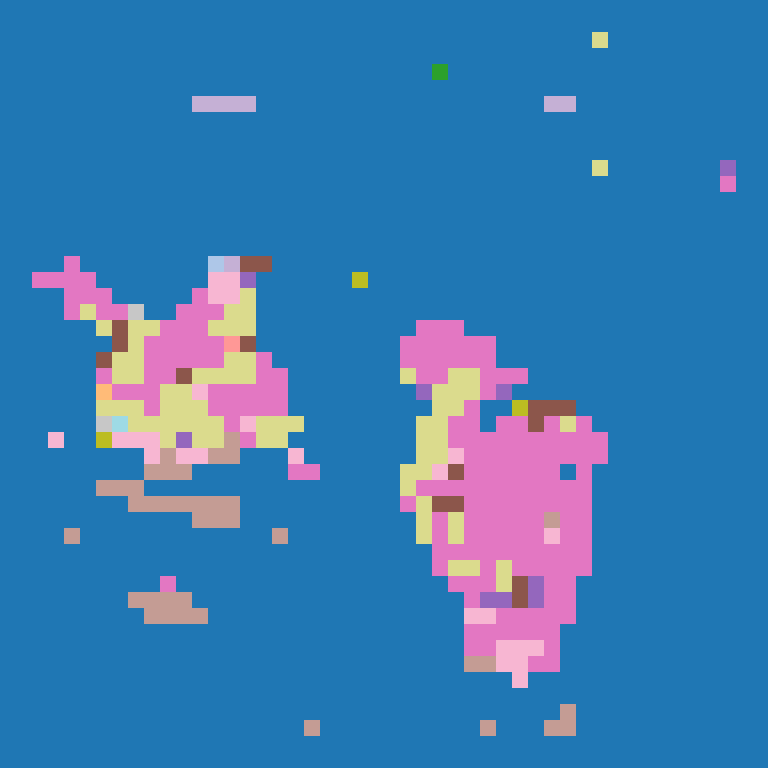}
        \end{tabular}
    \end{minipage}
    \hspace{0.001\linewidth} %
    \begin{minipage}{0.48\linewidth}
        \centering
        \renewcommand{\arraystretch}{0.7}
        \begin{tabular}{c c c}
            \footnotesize Input & \footnotesize Ours & \footnotesize SD 2.1 \\
            \includegraphics[width=0.32\linewidth]{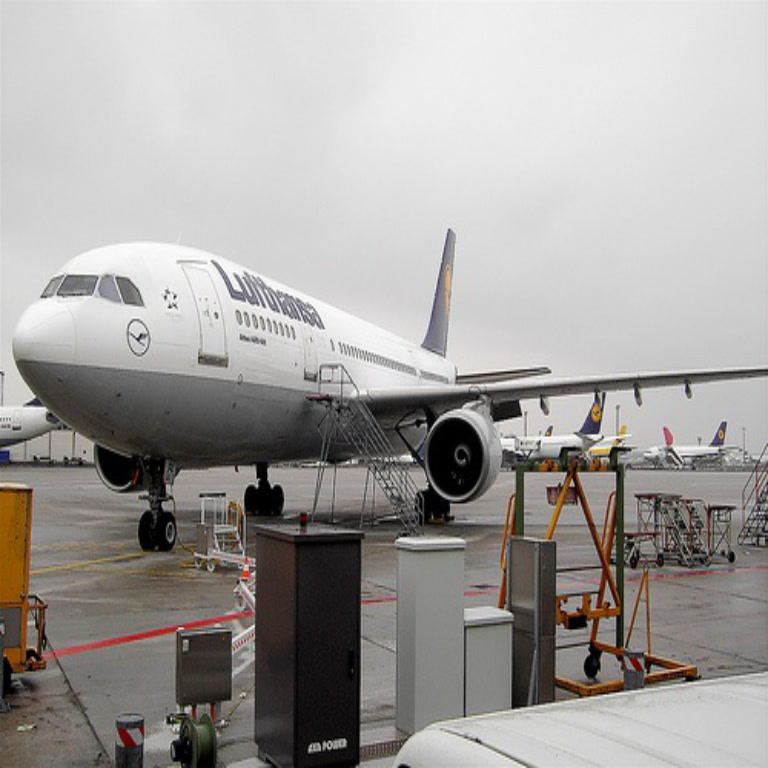} &
            \includegraphics[width=0.32\linewidth]{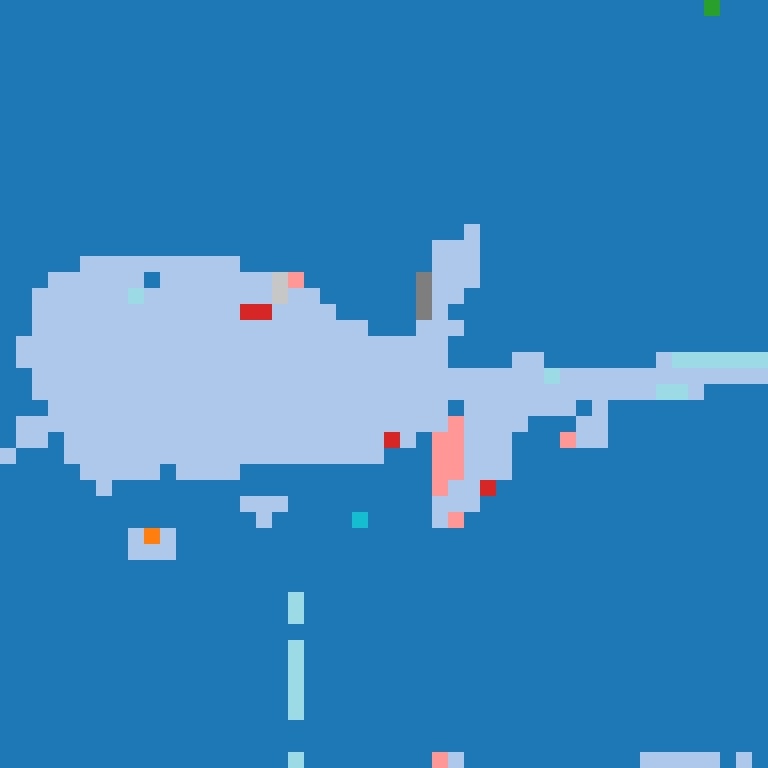} &
            \includegraphics[width=0.32\linewidth]{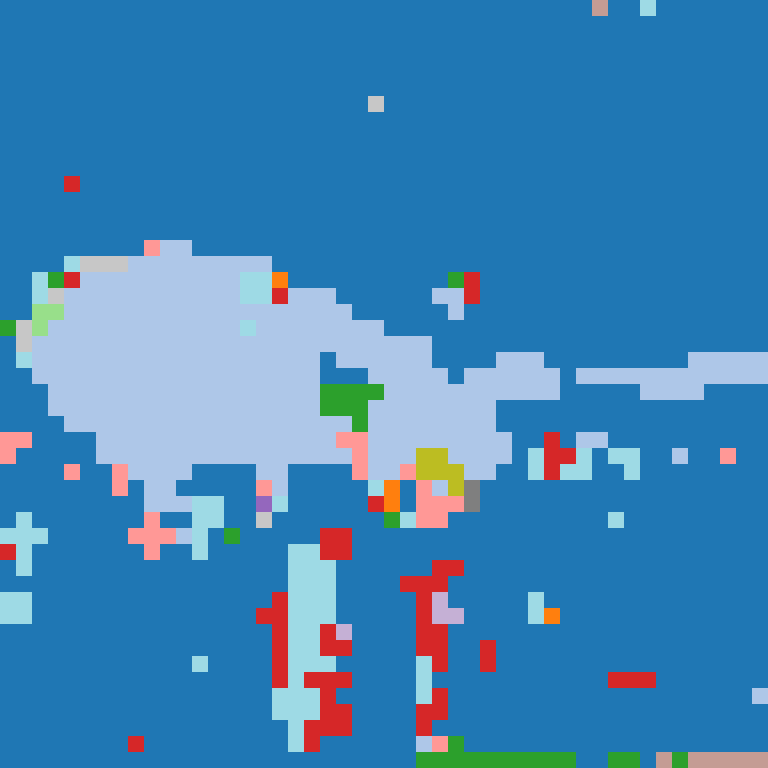} \\
            \includegraphics[width=0.32\linewidth]{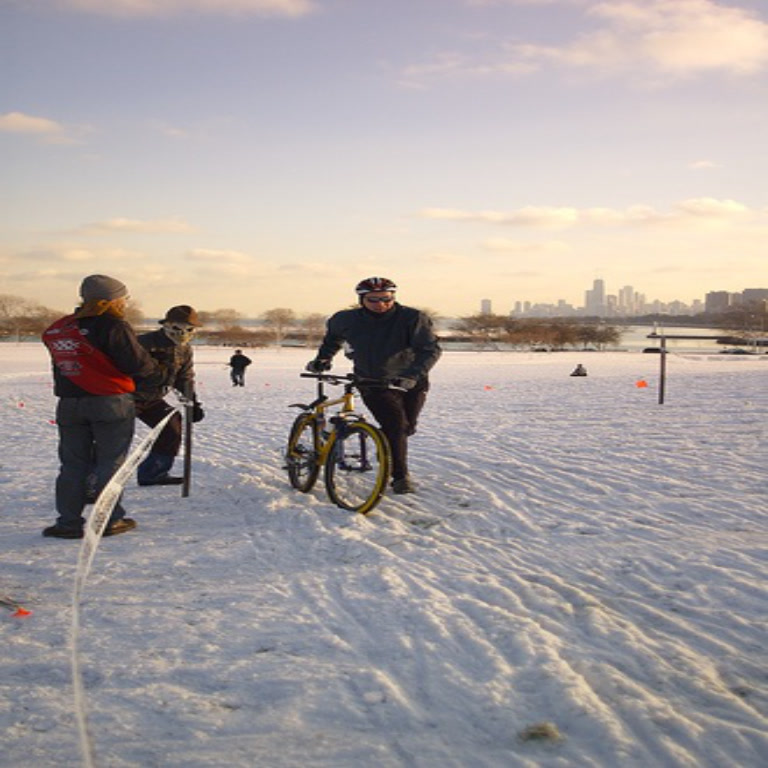} &
            \includegraphics[width=0.32\linewidth]{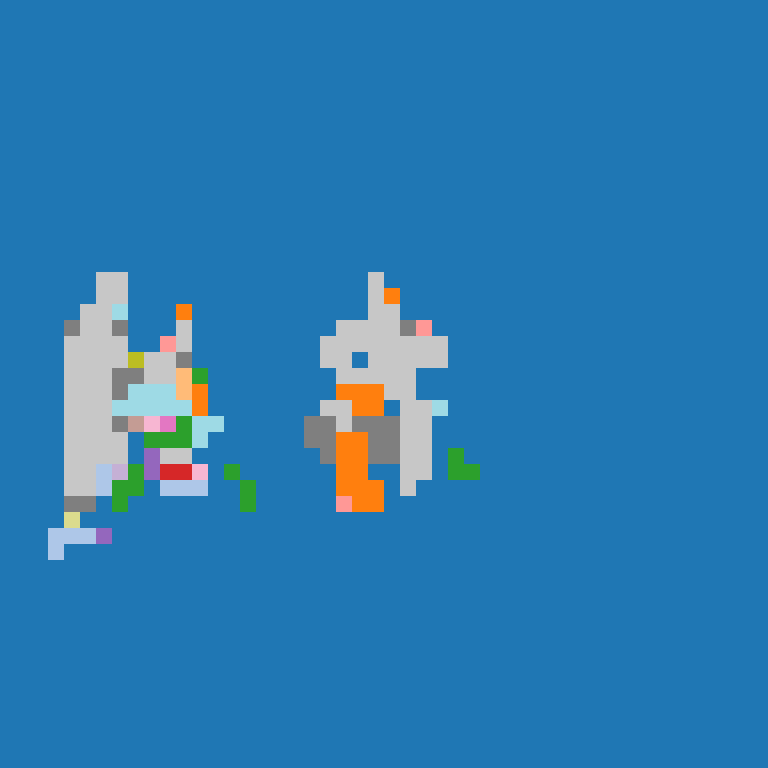} &
            \includegraphics[width=0.32\linewidth]{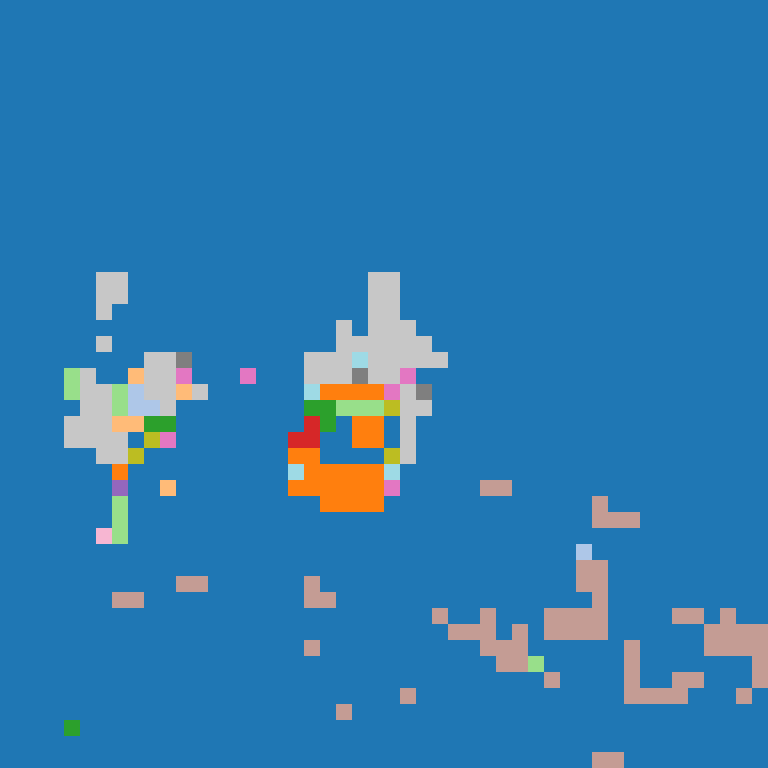} \\
            \includegraphics[width=0.32\linewidth]{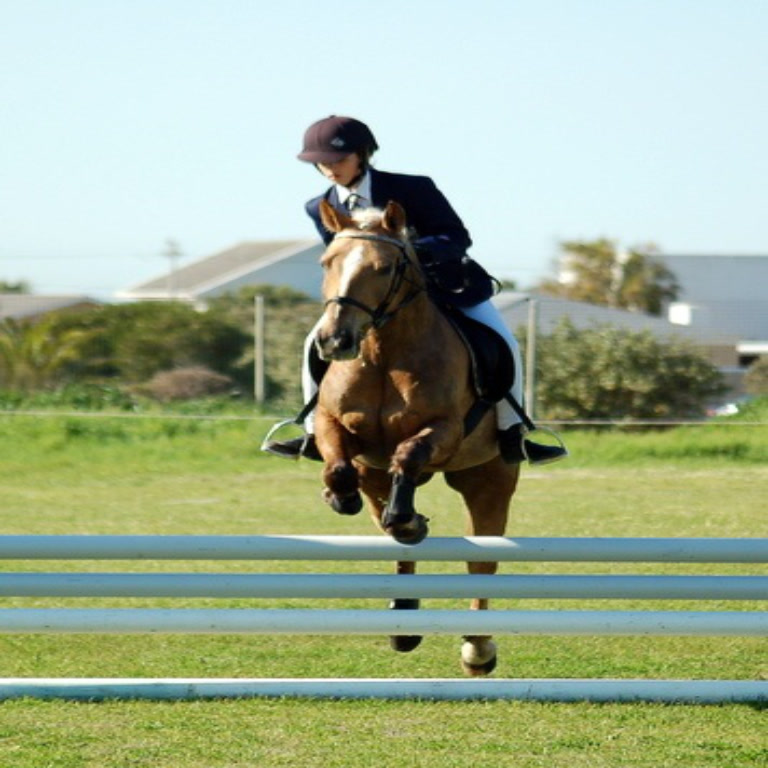} &
            \includegraphics[width=0.32\linewidth]{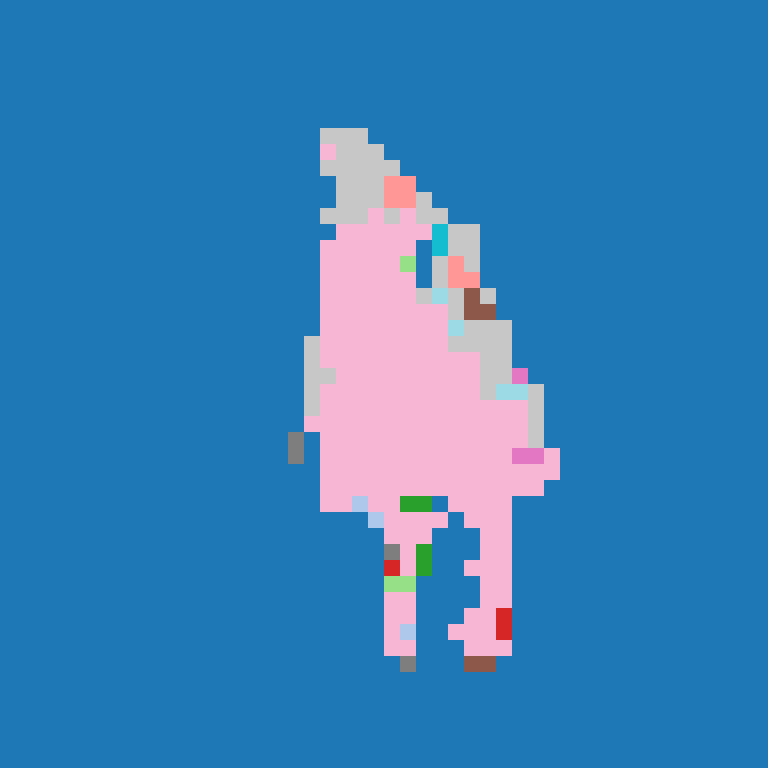} &
            \includegraphics[width=0.32\linewidth]{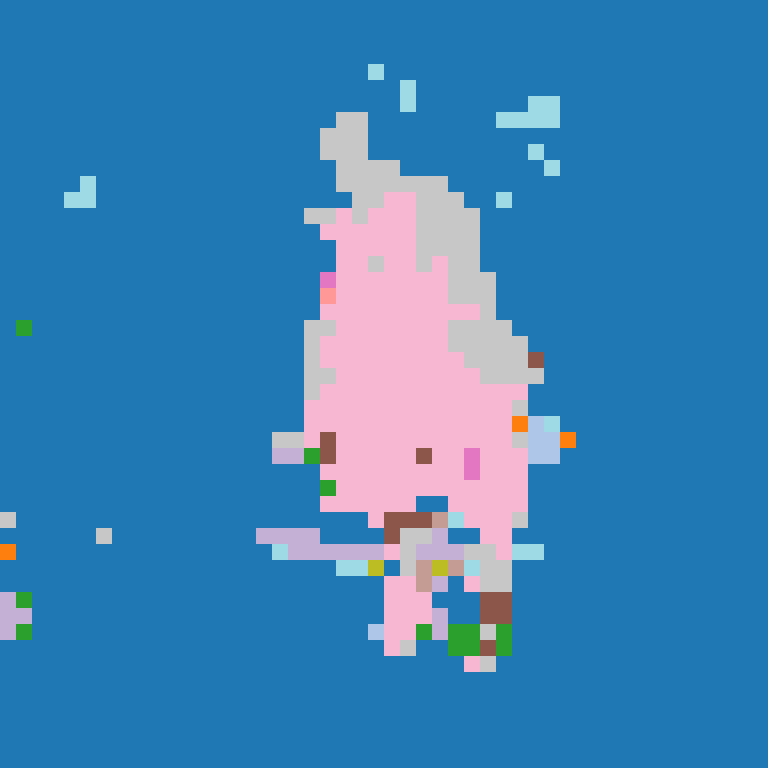}
        \end{tabular}
    \end{minipage}
    \caption{Qualitative results for semantic segmentation from diffusion features on Pascal VOC \cite{Everingham10}. Standard SD features use $t=100$ as the timestep, which we found to perform best quantitatively (cf.~\cref{fig:sem_seg_quantitative}). Note how the CleanDIFT segmentation maps are far less noisy than those of the standard diffusion features.}
    \label{fig:sem_seg}
\end{figure}
}

\begin{figure}[t]
    \centering
    \scalebox{.6}{\includegraphics[]{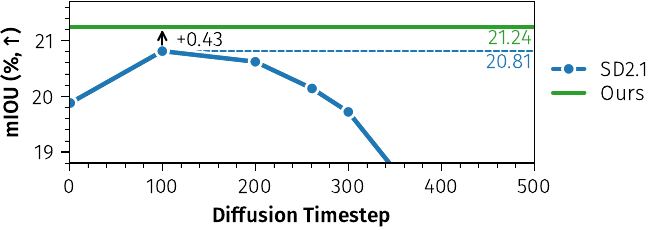}}
    \caption{Performance on semantic segmentation using linear probes. Our clean features outperform the noisy diffusion features for the best noising timestep $t$. Semantic segmentation performance of a standard diffusion model heavily depends on the used noising timestep. Unlike for semantic correspondence matching, the optimal $t$ value appears to be around $t=100$.\vspace{-4mm}}
    \label{fig:sem_seg_quantitative}
\end{figure}

\subsection{Classification}
\label{subsec:classification}
\begin{figure}[t]
    \centering
    \scalebox{.6}{\includegraphics[]{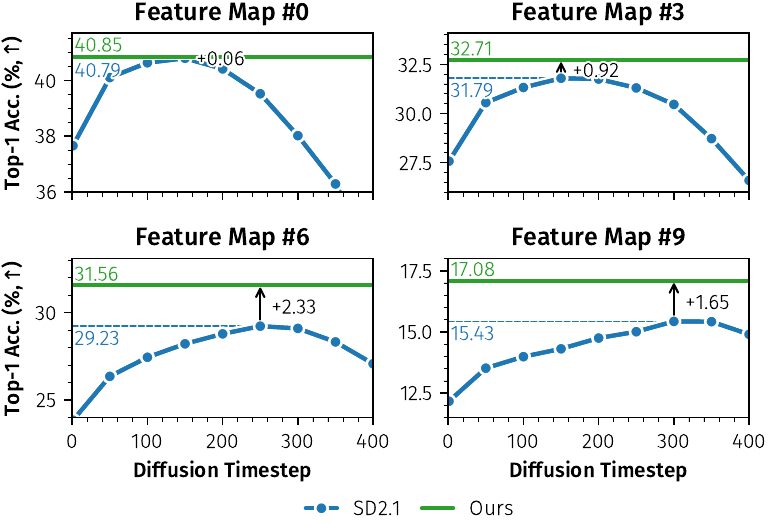}}
    \vspace{-1.5mm}
    \caption{Classification performance on ImageNet1k~\cite{deng_image_2009}, using kNN classifier with $k=10$ and cosine similarity as the distance metric. We sweep over different timesteps and feature maps. We find that the feature map with the lowest spatial resolution (feature map \#0) yields the highest classification accuracy. \vspace{-4mm}}
    \label{fig:class_quantitative}
\end{figure}

To assess the impact of our method on non-spatial tasks, we evaluate classification performance using pooled features. Pooling mitigates the influence of localized noise, so we anticipate classification performance to remain on par with standard diffusion features unless our setup introduces detrimental effects. We perform k-Nearest Neighbor (kNN) classification with $k = 10$ on ImageNet1k~\cite{deng_image_2009}, using SD 1.5 as the diffusion backbone. We sweep across feature maps and timesteps $t$ for the base model, with results presented in~\cref{fig:class_quantitative}. Our analysis shows that the feature map with the lowest spatial resolution, i.e., feature map \#0 (see~\cref{fig:architecture-details}), achieves the highest classification accuracy. Furthermore, the optimal timestep $t$ for the base model varies between feature maps. For the best-performing feature map, $t=100$ yields the highest classification accuracy.

Importantly, CleanDIFT features slightly outperform the standard diffusion features even when using an optimal timestep $t$ for the base model showing that it does not introduce any detrimental effects.

\subsection{Ablation Studies}
\label{subsec:ablation}

For simplicity, we perform our ablation studies using DIFT \cite{tang_emergent_2023} and evaluate the performance for unsupervised zero-shot semantic correspondence matching on a subset of the SPair71k~\cite{min2019spair} test split.

\vspace{-2mm}
\paragraph{Training Objective} During training, we maximize the cosine similarity between projected outputs of our feature extraction model and standard diffusion features to align them. To investigate the influence of the employed similarity metric, we compare feature extraction models trained on three different alignment objectives commonly used in similar contexts: mean absolute error ($L_1$), mean squared error ($L_2$), and cosine similarity. Quantitative results are provided in~\cref{tab:ablation}. While all objectives result in feature extraction models that outperform standard diffusion features, cosine similarity consistently performs the best across the alignment objectives by a significant margin, confirming our choice of similarity metric.

\begin{table}[h]

\centering
\begin{tabular}{@{\hskip .5em}lcccc@{\hskip .5em}}
\toprule
\multirow{2}{*}[-3pt]{Objective} & \multirow{2}{*}[-3pt]{\shortstack{Projection\\Heads}} & \multicolumn{2}{c}{{PCK@$\alpha\ (\uparrow)$}} \\ \cmidrule(lr){3-4}
                                     &                                    & $\alpha_\mathrm{img} = 0.1$ & $\alpha_\mathrm{bbox} = 0.1$ \\
\midrule
\multirow{2}{*}{Cosine Sim.} & \cmark & \textbf{68.32} & \textbf{61.43} \\
                        & {\xmark} & \underline{68.16} & \underline{61.29} \\
\cmidrule{2-4}
\multirow{2}{*}{$L_2$}     & \cmark & 66.23 & 59.13 \\
                        & {\xmark} & 66.49 & 59.43 \\
\cmidrule{2-4}
\multirow{2}{*}{$L_1$}     & \cmark & 66.91 & 60.00 \\
                        & {\xmark} & 66.87 & 59.91 \\
\midrule
\midrule
SD 2.1                 & -                      & 63.41 & 55.92 \\
\bottomrule
\end{tabular}
\caption{Ablation Study Results. We evaluate the feature extraction models' performance for zero-shot semantic correspondence matching on the SPair71k test split. PCK is aggregated per point. \vspace{-4mm}}
\label{tab:ablation}
\end{table}

\vspace{-2mm}
\paragraph{Projection Heads} We investigate the influence of our proposed projection heads that are used to project CleanDIFT features onto standard diffusion features. The alignment of feature extraction model and diffusion model is determined by the utilized similarity metric and the projection heads. Therefore, we evaluate our feature extraction model's performance with and without the projection heads in combination with all three similarity metrics. An overview of the comparison is given in~\cref{tab:ablation}. In our main configuration that uses cosine similarity, using the projection heads yields slight performance improvements of $0.24$ percentage points for $\text{PCK}_{\text{img}}$ and $0.06$ percentage points for $\text{PCK}_{\text{bbox}}$ compared to fine-tuning without the projection heads. As the projection heads are typically not used for inference, they add computational overhead only during the lightweight fine-tuning. Therefore, we argue that it is worthwhile to include them and leverage the small performance gain. Additionally, they can be reused to efficiently obtain feature maps for specific timesteps.

\section{Conclusion}

In this paper, we introduced CleanDIFT, a novel approach for extracting diffusion features. CleanDIFT produces noise-free, timestep-independent, general-purpose diffusion features by consolidating timestep-dependent representations from a pre-trained diffusion backbone into a unified feature representation. We achieve this alignment between our feature extraction model and the pre-trained diffusion backbone through a lightweight fine-tuning procedure that takes approximately 30 minutes on a single A100 GPU. Operating directly on clean images, our method eliminates the information loss associated with adding noise to input images. Furthermore, CleanDIFT removes the requirement for tuning timesteps for each downstream task and avoids the computational overhead of ensembling over noise levels or timesteps. Instead, our method efficiently extracts features with just a single forward pass at inference time, substantially reducing inference costs compared to methods relying on ensembling or inversion. Extensive evaluations of CleanDIFT across diverse downstream tasks demonstrate significant performance improvements over conventional diffusion features.

    \section*{Acknowledgments}
    This project has been supported by the German Federal Ministry for Economic Affairs and Energy within the project ``NXT GEN AI METHODS – Generative Methoden für Perzeption, Prädiktion und Planung'', the project ``GeniusRobot'' (01IS24083), funded by the Federal Ministry of Education and Research (BMBF), the bidt project KLIMA-MEMES, and the German Research Foundation (DFG) project 421703927. The authors gratefully acknowledge the Gauss Center for Supercomputing for providing compute through the NIC on JUWELS at JSC and the HPC resources supplied by the Erlangen National High Performance Computing Center (NHR@FAU funded by DFG project 440719683) under the NHR project JA-22883.
    
    Further, we would like to thank Owen Vincent for continuous technical support.

{
    \small
    \bibliographystyle{ieeenat_fullname}
    \bibliography{main}
}

\clearpage

\maketitlesupplementary

\setcounter{section}{0}
\renewcommand{\thesection}{\Alph{section}}

\begin{table*}[t]
\centering
\resizebox{\textwidth}{!}{%
\begin{tabular}{@{\hskip .5em}l@{\hskip .5em}ccccccccccccccccccc!{\vrule width 0.8pt}l@{\hskip .5em}}
\toprule
\multirow{2}{*}[-3pt]{Method} & \multirow{2}{*}[-3pt]{\shortstack{Our \\ Features}} & \multicolumn{19}{c}{PCK@$\alpha_\mathrm{bbox}$ = 0.1 per category ($\uparrow$)} \\ 
\cmidrule(r){3-20}
 &  & Aero & Bike & Bird & Boat & Bottle & Bus & Car & Cat & \textcolor{blue}{Chair} & Cow & Dog & Horse & Motor & Person & \textcolor{blue}{Plant} & Sheep & Train & \textcolor{blue}{TV} & \makecell[c]{\textbf{All}} \\ 
\midrule
\multirow{2}{*}{DIFT \cite{tang_emergent_2023}} 
 & {\color{ourred}\xmark}  & 63.41 & 55.10 & 80.40 & 34.55 & 46.15 & 52.26 & 48.02 & 75.86 & {39.46} & 75.57 & 55.00 & 61.71 & 53.32 & 46.53 & {56.36} & 57.68 & 71.30 & {\underline{63.63}} & \ 59.57 \\ 
 & {\color{ourgreen}\cmark}  & \cellcolor{ourwhite}63.72 & \cellcolor{ourwhite}55.90 & \cellcolor{ourwhite}80.50 & \cellcolor{ourwhite}35.40 & \cellcolor{ourwhite}49.36 & \cellcolor{ourwhite}53.46 & \cellcolor{ourwhite}48.08 & \cellcolor{ourwhite}75.78 & \cellcolor{ourwhite}{43.10} & \cellcolor{ourwhite}76.20 & \cellcolor{ourwhite}55.69 & \cellcolor{ourwhite}61.01 & \cellcolor{ourwhite}54.17 & \cellcolor{ourwhite}49.14 & \cellcolor{ourwhite}{\underline{62.56}} & \cellcolor{ourwhite}58.37 & \cellcolor{ourwhite}74.63 & \cellcolor{ourwhite}{\textbf{71.54}} & \ \cellcolor{ourwhite}61.43\textsubscript{\color{ourgreen}$\blacktriangle$1.86} \\ 
\multirow{2}{*}{A Tale of Two Features \cite{zhang_tale_2023}} 
 & {\color{ourred}\xmark}   & 71.26 & 62.23 & 87.01 & 37.24 & 53.78 & 54.32 & 51.20 & 78.61 & 46.50 & 78.93 & 64.43 & 69.47 & 62.23 & 69.27 & {59.28} & 68.03 & {65.40} & {53.81} & \ 63.73 \\ 
 & {\color{ourgreen}\cmark} & \cellcolor{ourwhite}71.12 & \cellcolor{ourwhite}62.70 & \cellcolor{ourwhite}87.42 & \cellcolor{ourwhite}38.33 & \cellcolor{ourwhite}\textbf{54.78} & \cellcolor{ourwhite}54.67 & \cellcolor{ourwhite}51.20 & \cellcolor{ourwhite}78.52 & \cellcolor{ourwhite}47.86 & \cellcolor{ourwhite}79.38 & \cellcolor{ourwhite}64.88 & \cellcolor{ourwhite}69.18 & \cellcolor{ourwhite}62.61 & \cellcolor{ourwhite}69.72 & \cellcolor{ourwhite}{\textbf{62.82}} & \cellcolor{ourwhite}68.87 & \cellcolor{ourwhite}{67.51} & \cellcolor{ourwhite}{59.04} & \ \cellcolor{ourwhite}64.81\textsubscript{\color{ourgreen}$\blacktriangle$1.08} \\ 
\multirow{2}{*}{Telling Left from Right \cite{zhang_telling_2024}} 
 & {\color{ourred}\xmark}  & \textbf{78.14} & \textbf{66.37} & \textbf{89.60} & \underline{43.74} & 53.29 & \underline{66.61} & \underline{59.94} & \textbf{82.66} & {\underline{51.75}} & \textbf{82.79} & \textbf{68.95} & \underline{74.91} & \underline{65.84} & \textbf{71.67} & {57.71} & \textbf{72.24} & \underline{83.46} & {49.66} & \ \underline{68.64} \\ 
 & {\color{ourgreen}\cmark}  & \cellcolor{ourwhite}\underline{77.17} & \cellcolor{ourwhite}\underline{65.65} & \cellcolor{ourwhite}\underline{89.58} & \cellcolor{ourwhite}\textbf{44.24} & \cellcolor{ourwhite}\underline{54.27} & \cellcolor{ourwhite}\textbf{67.24} & \cellcolor{ourwhite}\textbf{60.63} & \cellcolor{ourwhite}\underline{82.33} & \cellcolor{ourwhite}{\textbf{56.57}} & \cellcolor{ourwhite}\underline{82.53} & \cellcolor{ourwhite}\underline{68.37} & \cellcolor{ourwhite}\textbf{75.91} & \cellcolor{ourwhite}\textbf{65.99} & \cellcolor{ourwhite}\underline{71.37} & \cellcolor{ourwhite}{62.29} & \cellcolor{ourwhite}\underline{70.42} & \cellcolor{ourwhite}\textbf{84.58} & \cellcolor{ourwhite}{59.84} & \ \cellcolor{ourwhite}\textbf{69.99}\textsubscript{\color{ourgreen}$\blacktriangle$1.35} \\ 
\bottomrule
\end{tabular}%
}
\caption{Reproduced results for zero-shot unsupervised semantic correspondence matching, evaluated on SPair71k~\cite{min2019spair}. The three categories for which we observe the largest overall gains are marked in \textcolor{blue}{blue}. We report PCK@$\alpha=0.1$ with an error margin relative to bounding box sizes on the test split of SPair71k, aggregated per point and per category. We compare our reproductions against the papers' reported numbers in~\cref{tab:reported_vs_reproduced}}
\label{tab:correspondence-quantitative-per-cat}
\end{table*}

\begin{table}[t]
\centering
\adjustbox{max width=\linewidth}{
\begin{tabular}{@{\hskip .5em}l@{\hskip .5em}ccc@{\hskip .5em}}
\toprule
\multirow{2}{*}[-3pt]{Method} & \multirow{2}{*}[-3pt]{\shortstack{Eval\\Method}} & \multicolumn{2}{c}{PCK@$\alpha\ (\uparrow)$ } \\ \cmidrule(lr){3-4}
                        && $\alpha_\mathrm{img}$ = 0.1 & $\alpha_\mathrm{bbox}$ = 0.1   \\ 
\midrule
\multirow{2}{*}{DIFT \cite{tang_emergent_2023}}
 & reported &  --         & 59.50               \\
& \cellcolor{ourwhite}reproduced & \cellcolor{ourwhite}66.53  & \cellcolor{ourwhite}59.57          \\
\multirow{2}{*}{A Tale of Two Features \cite{zhang_tale_2023}} & reported & --          & 64.00              \\
& \cellcolor{ourwhite}reproduced & \cellcolor{ourwhite}72.31  & \cellcolor{ourwhite}63.73        \\
\multirow{2}{*}{Telling Left from Right \cite{zhang_telling_2024}} & reported  & --       & 69.60               \\
& \cellcolor{ourwhite}reproduced & \cellcolor{ourwhite}77.07        & \cellcolor{ourwhite}68.64           \\ 

\bottomrule
\end{tabular}
}
\caption{Reproduced vs reported numbers for zero-shot semantic correspondences, evaluated on SPair71k~\cite{min2019spair}. A Tale of Two Features~\cite{zhang_tale_2023} and Telling Left from Right~\cite{zhang_telling_2024} report higher PCK values than our reproduction because they utilize a conditioning mechanism on CLIP image embeddings from~\cite{xu2023open} that was fine-tuned for panoptic segmentation. As this task is related to semantic correspondence matching, we do not consider using this conditioning mechanism fair in comparison to other zero-shot approaches for semantic correspondences. Therefore, we exclude it from our reproductions.} 
\label{tab:reported_vs_reproduced}
\end{table}

\begin{figure}[t]
    \centering
    \includegraphics[width=\columnwidth]{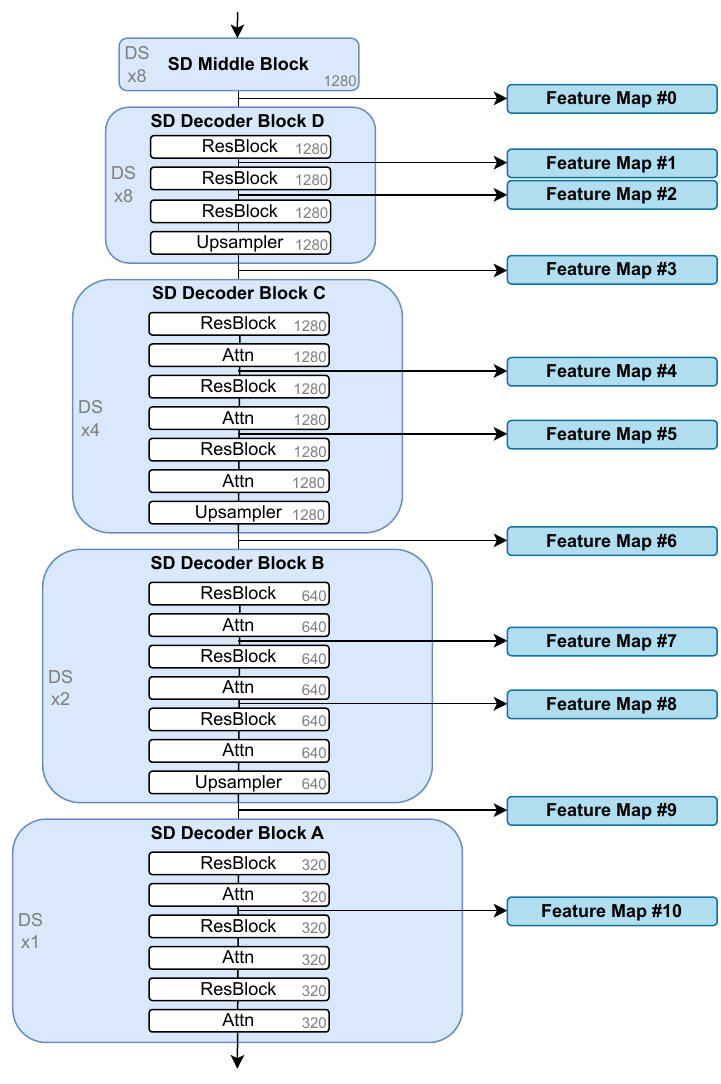}
    \caption{We align the feature maps between our feature extractor and the diffusion model at multiple stages within the network to enable the usage of multiple feature maps for downstream tasks. In total, we extract and align features at $K=11$ stages of the SD U-Net decoder. The downsampling factor for the different blocks is denoted as \textit{DS} and the channel dimension is shown on the right side of each block.}
    \label{fig:architecture-details}
\end{figure}

\section{Architecture Details}
\label{sec:arch-details}

An illustration of where exactly we extract and align feature maps is provided in~\cref{fig:architecture-details}. The decoder architecture is identical for SD 1.5 and SD 2.1, therefore~\cref{fig:architecture-details} applies to both models. DIFT~\cite{tang_emergent_2023} extracts feature map \#6. A Tale of Two Features~\cite{zhang_tale_2023} and Telling Left from Right~\cite{zhang_telling_2024} both extract feature maps \#2, \#6, and \#8.

\begin{figure}[t]
    \centering
    \includegraphics[width=\columnwidth]{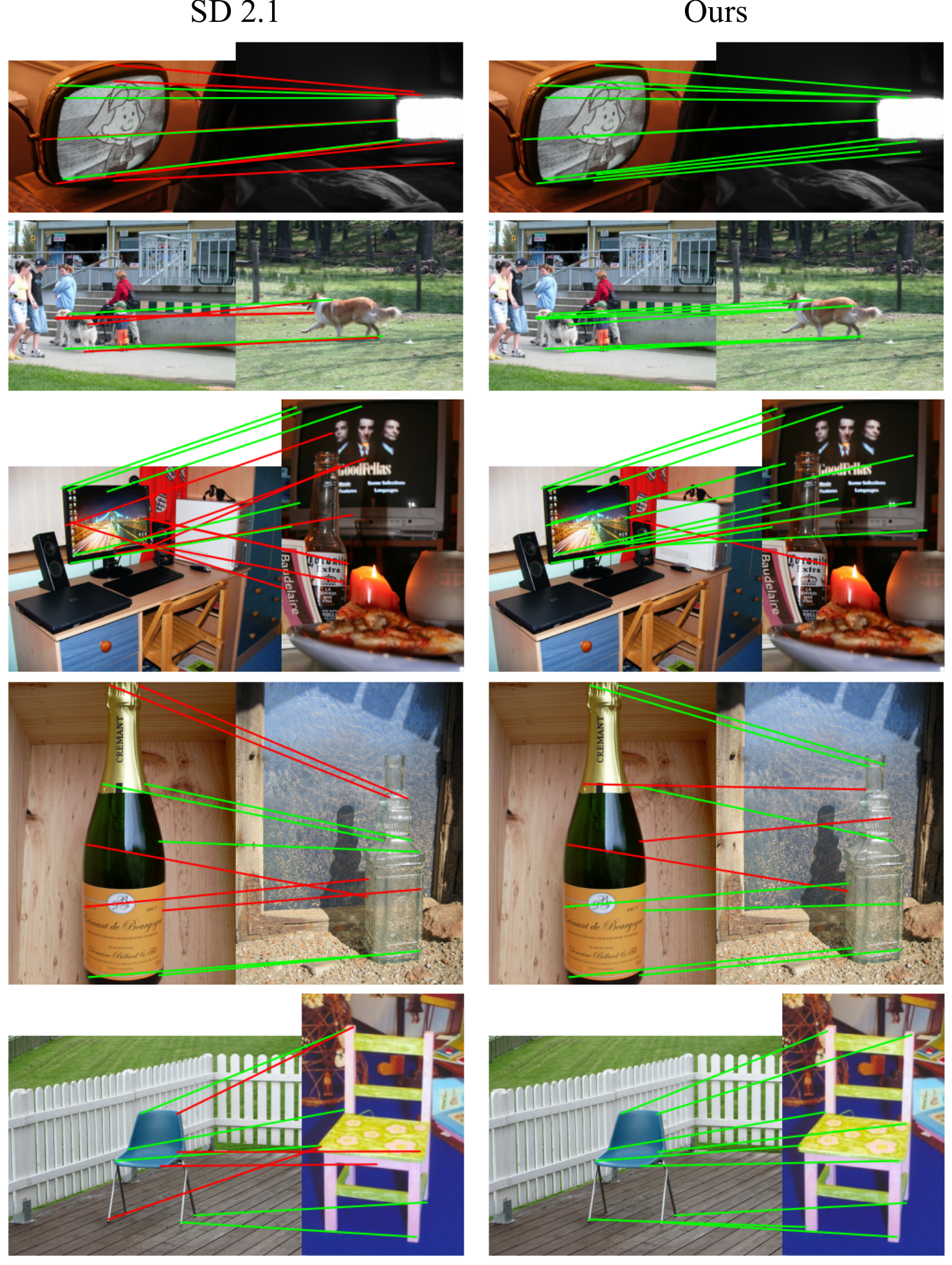}
    \caption{Additional qualitative results for semantic correspondence matching using DIFT~\cite{tang_emergent_2023} with the standard SD 2.1 ($t=261$) and our CleanDIFT features. Our clean features show significantly less incorrect matches than the standard diffusion features, especially along texture-less edges.}
    \label{fig:supp-correspondence-examples}
\end{figure}

\section{Additional Quantitative Evaluations}
\label{sec:add-quant-eval}

\paragraph{Unsupervised Semantic Correspondence}

In~\cref{tab:correspondence-quantitative-per-cat}, we provide an extended version of~\suppvspreprint{Tab. 1}{\cref{tab:correspondence-quantitative}}, in which we report the PCK metric per category of the SPair71k dataset~\cite{min2019spair}. The categories for which we observe the largest gain when comparing our CleanDIFT features to standard diffusion features are \textit{TV}, \textit{Plant}, and \textit{Chair}. These classes are characterized by long, texture-less edges: the bezel of a TV monitor, the pot of a plant, and the legs of a chair. Supporting this observation, the performance gain for samples from the \textit{Plant} class mostly comes from keypoints not on the plant itself but on the pot of the plant. This is further illustrated in~\suppvspreprint{Figure 6}{\cref{fig:correspondence-examples}}. We conclude that our CleanDIFT features are particularly effective for matching corresponding keypoints located along texture-less edges.

\paragraph{Supervised Semantic Correspondence}
We test our method in the supervised setting of TLFR~\cite{zhang_telling_2024} that uses an aggregation network~\cite{luo_diffusion_2024, zhang_tale_2023} to fuse feature maps. In this setting, our features achieve a \pck{} per image of $83.9$, outperforming SD2.1 features (83.2). This shows that even in a supervised setting, our features provide additional information that cannot be extracted by additional supervised training.   

\paragraph{Distilled Text Conditioning}
In our standard setting, we train CleanDIFT with image-text pairs since that is what the diffusion model expects as input. As a result, the model also requires a fitting text prompt for optimal feature performance at inference time. Here, we experiment with distilling the text conditioning during our fine-tuning to directly yield optimal features without the need for a prompt. To that end, we train a CleanDIFT version that depends neither on explicit nor implicit captions~\cite{zhang_tale_2023,zhang_telling_2024} by distilling the text conditioning directly into its features. This results in a $1.2$ \pck{} gain over DIFT SD2.1 features with 8x ensembling and no text prompt when extracting features. Compared to simply training CleanDIFT without captions and no text conditioning during inference, the distillation improves performance by 0.4 \pck{}.

\section{Additional Qualitative Samples}
\label{sec:supp-add-qual}

We provide additional qualitative samples for semantic correspondence matching in~\cref{fig:supp-correspondence-examples} and for semantic segmentation in~\cref{fig:add_sem_seg_qualitative}. \cref{fig:pca_qualitative} shows a PCA visualization of our features revealing that they are less noisy than standard features.

\section{Depth Probes Analysis}
\label{supp:depth}
We show a more thorough analysis of the depth probe experiment presented in \suppvspreprint{Sec. 4.3}{~\cref{subsec:depth}}. We show the full set of linear probes for depth prediction on our projected features, i.e. the outputs of the projection heads for different timesteps. A comparison of the performance across timesteps is provided in~\cref{fig:depth-over-timesteps}. We observe that the performance of linear probes trained on the projected features decreases for large timesteps, albeit significantly less severe than for standard diffusion features due to the absence of noise. The best performance on projected features is achieved at timestep $t=499$, while the best performance for standard diffusion features is achieved at timestep $t=299$.

{
\setlength{\tabcolsep}{0.005\linewidth}
\begin{figure}[t]
    \centering
    \begin{minipage}{0.48\linewidth}
        \centering
        \renewcommand{\arraystretch}{0.7}
        \begin{tabular}{@{\hskip .5em}c c c@{\hskip .5em}}
            \small Input & \small Ours & \small SD 2.1 \\
            \includegraphics[width=0.32\linewidth]{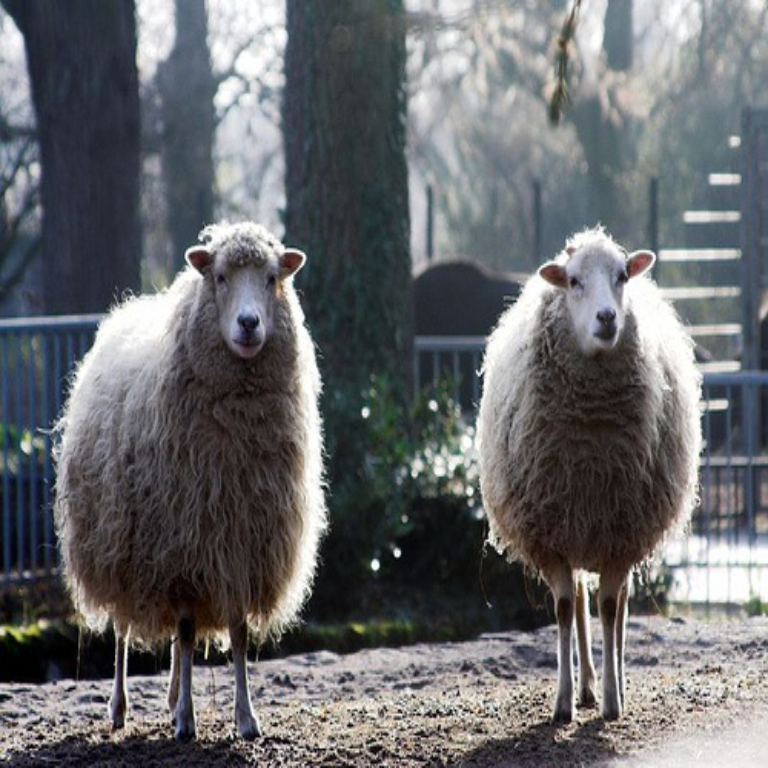} &
            \includegraphics[width=0.32\linewidth]{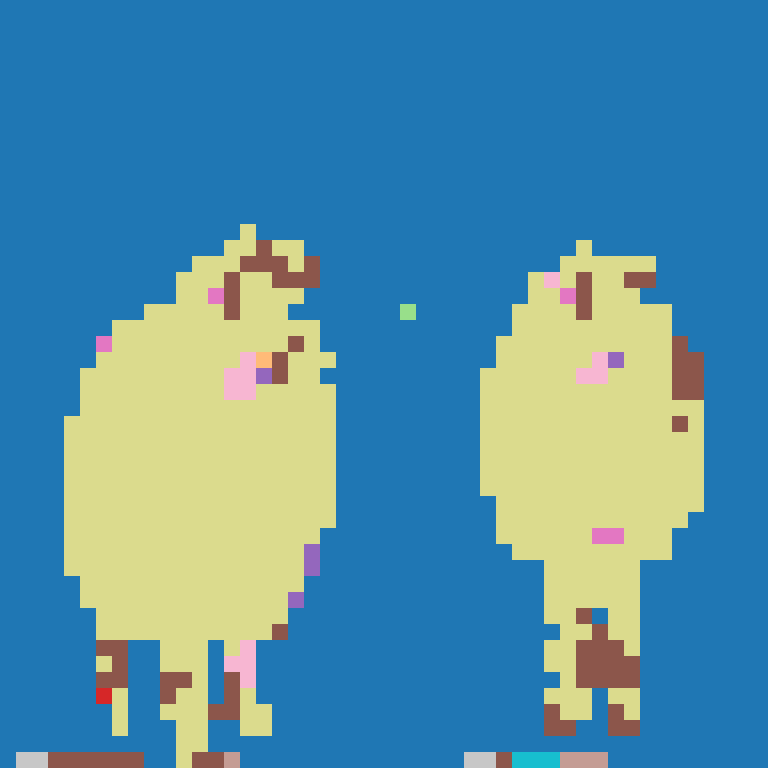} &
            \includegraphics[width=0.32\linewidth]{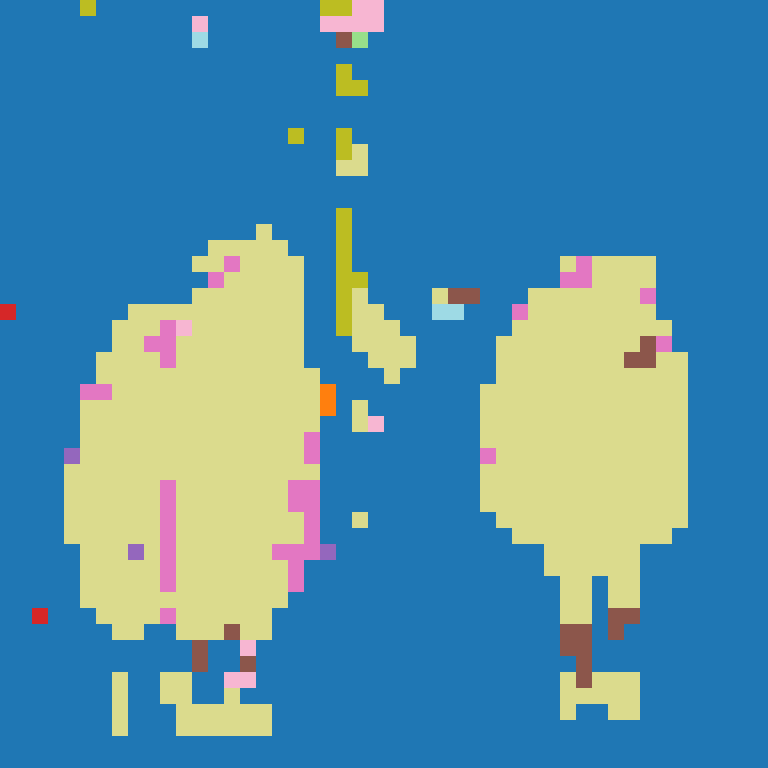} \\
            \includegraphics[width=0.32\linewidth]{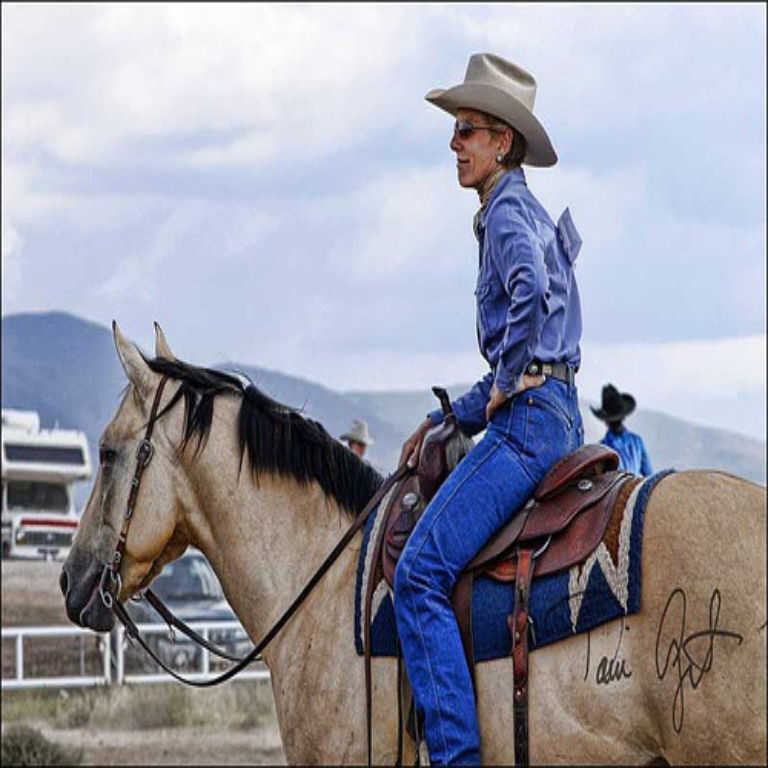} &
            \includegraphics[width=0.32\linewidth]{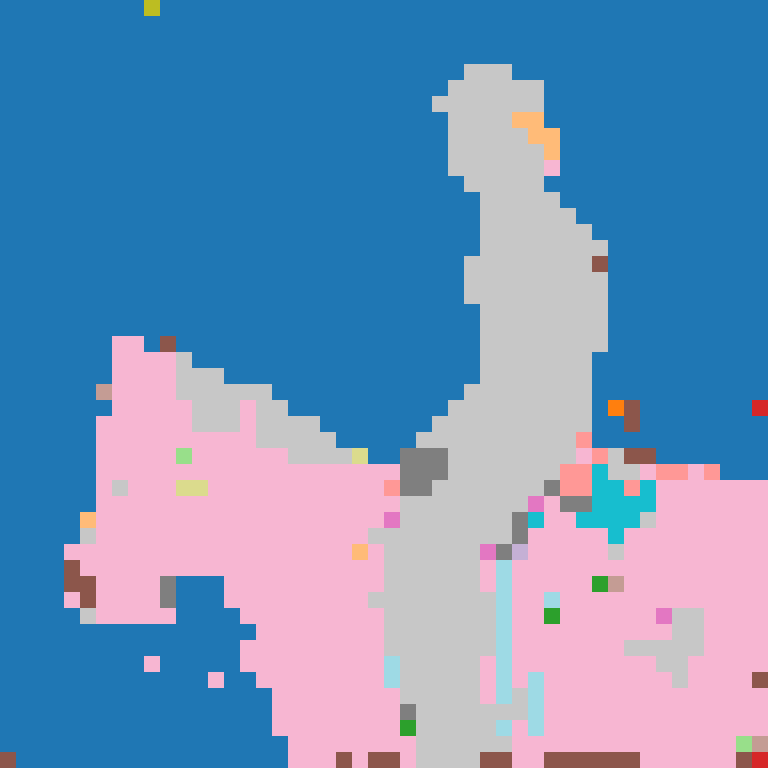} &
            \includegraphics[width=0.32\linewidth]{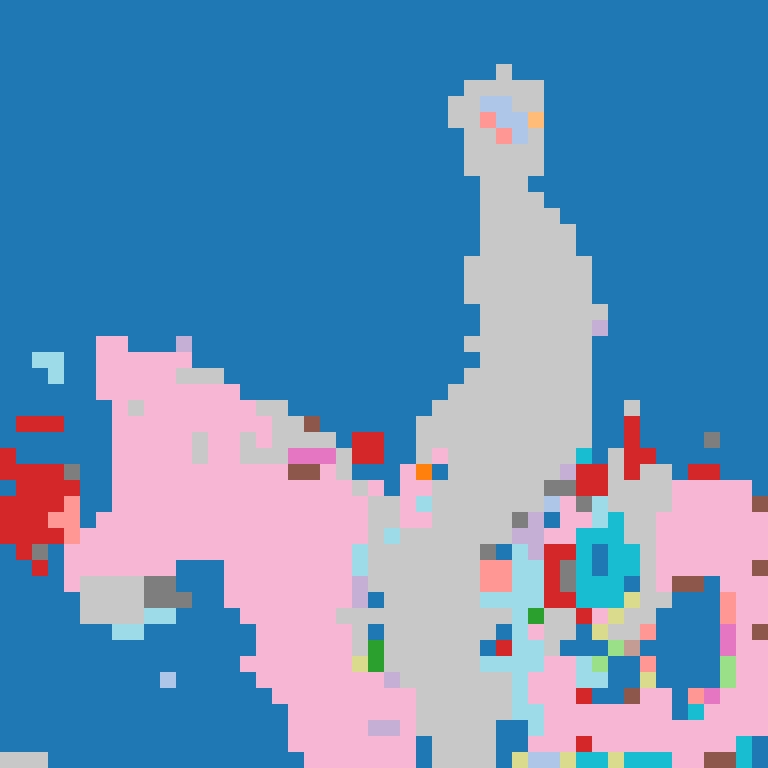} \\
            \includegraphics[width=0.32\linewidth]{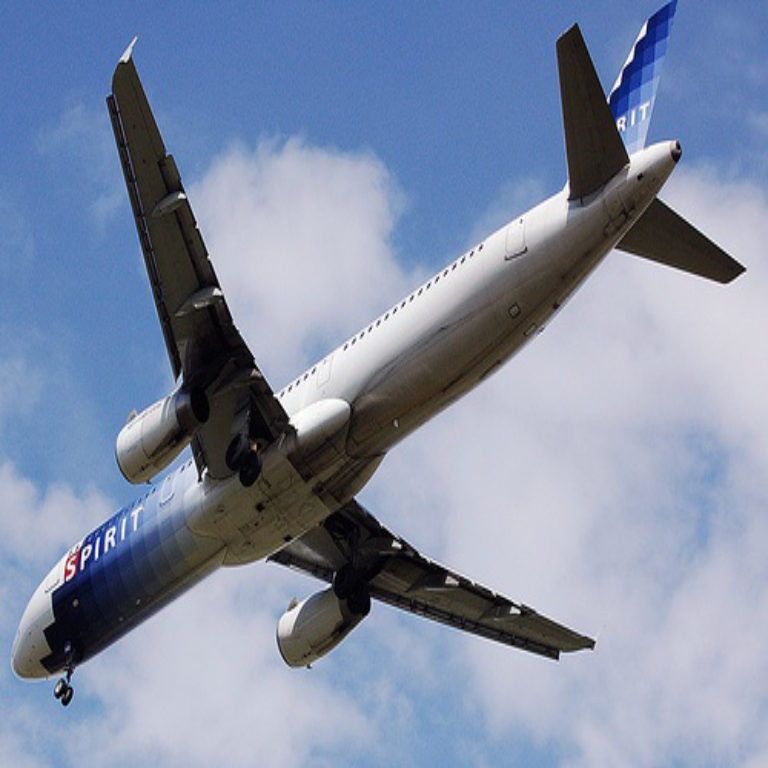} &
            \includegraphics[width=0.32\linewidth]{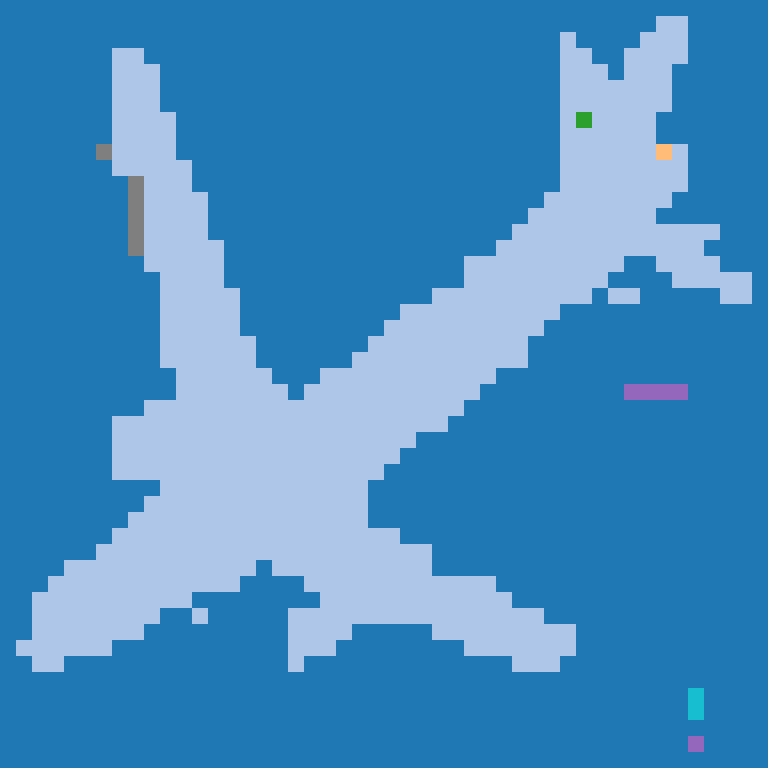} &
            \includegraphics[width=0.32\linewidth]{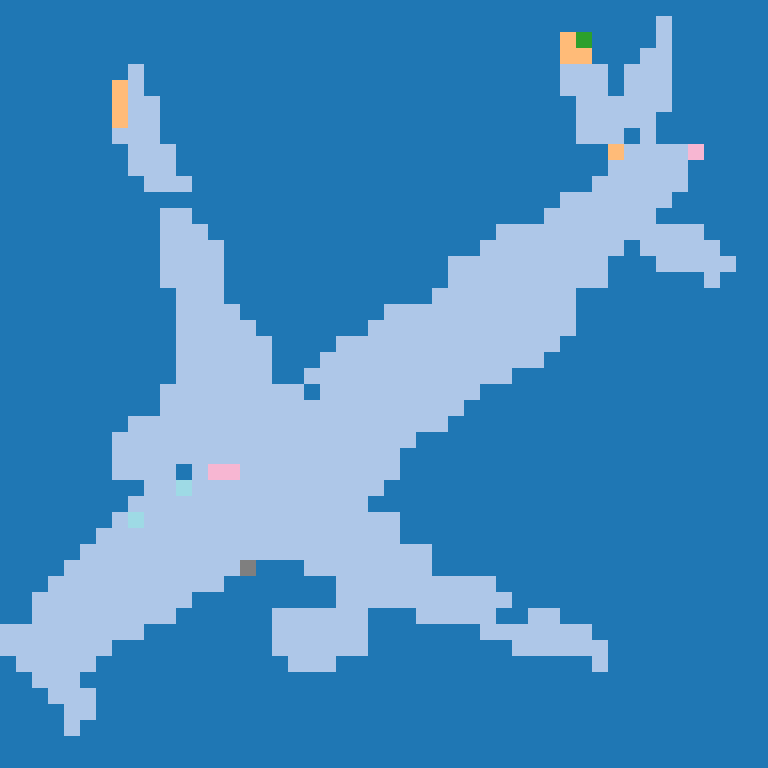}
        \end{tabular}
    \end{minipage}
    \hspace{0.001\linewidth} %
    \begin{minipage}{0.48\linewidth}
        \centering
        \renewcommand{\arraystretch}{0.7}
        \begin{tabular}{c c c}
            \small Input & \small Ours & \small SD 2.1 \\
            \includegraphics[width=0.32\linewidth]{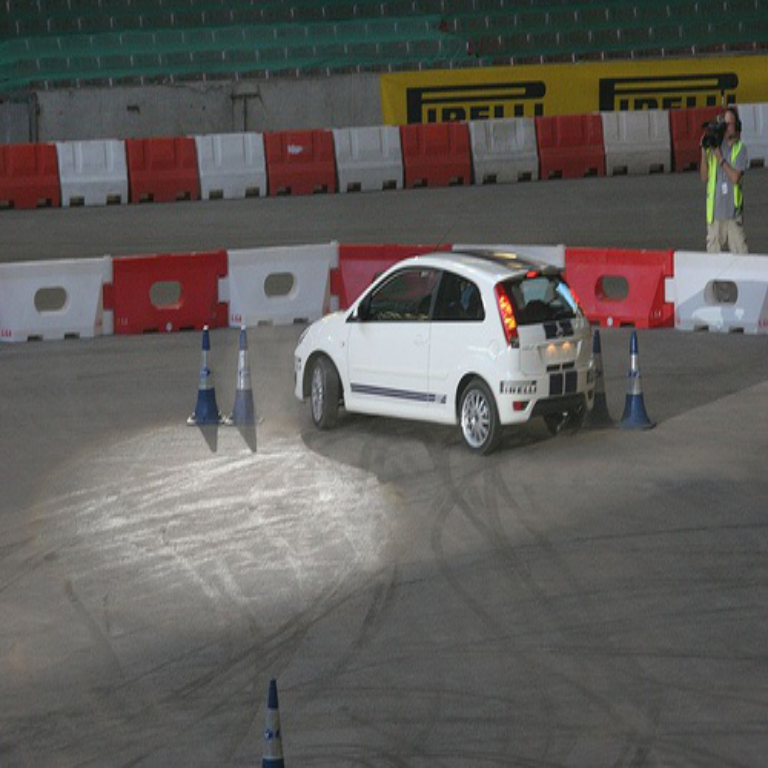} &
            \includegraphics[width=0.32\linewidth]{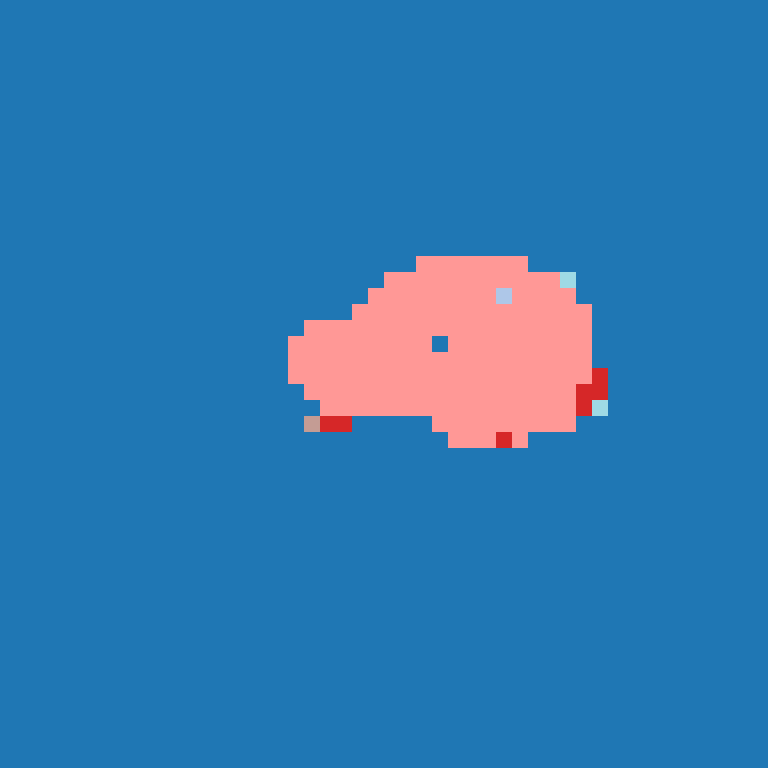} &
            \includegraphics[width=0.32\linewidth]{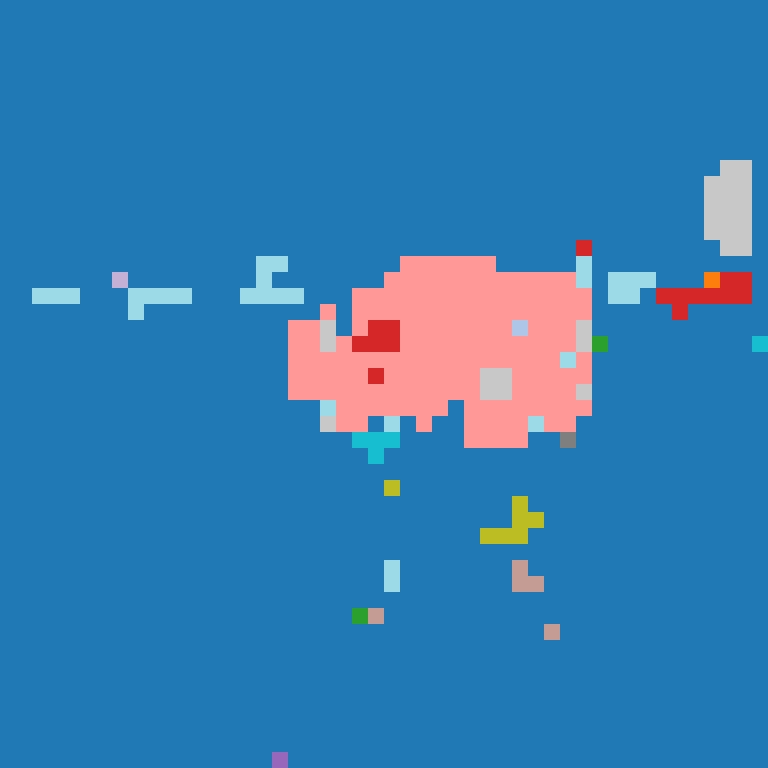} \\
            \includegraphics[width=0.32\linewidth]{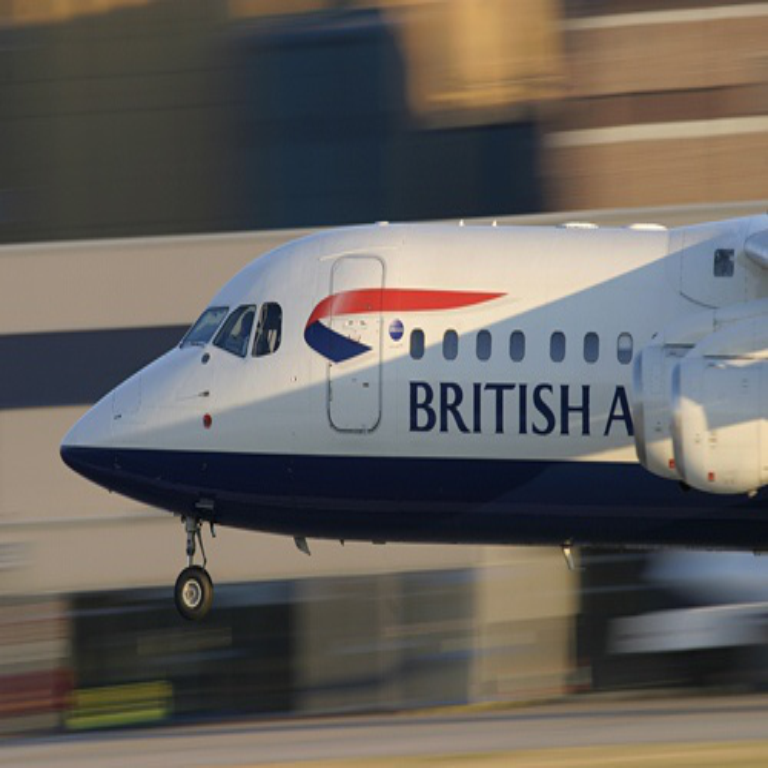} &
            \includegraphics[width=0.32\linewidth]{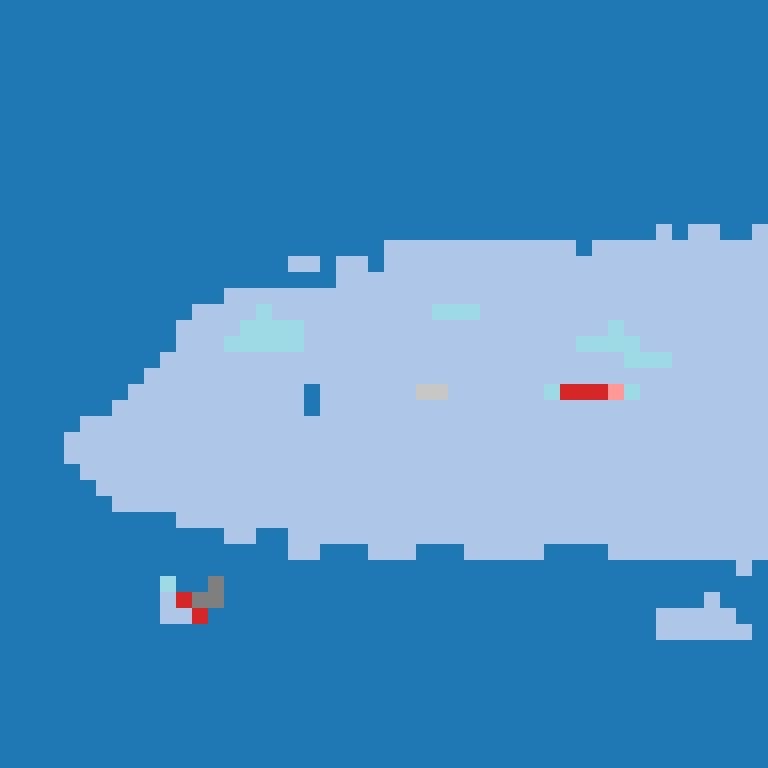} &
            \includegraphics[width=0.32\linewidth]{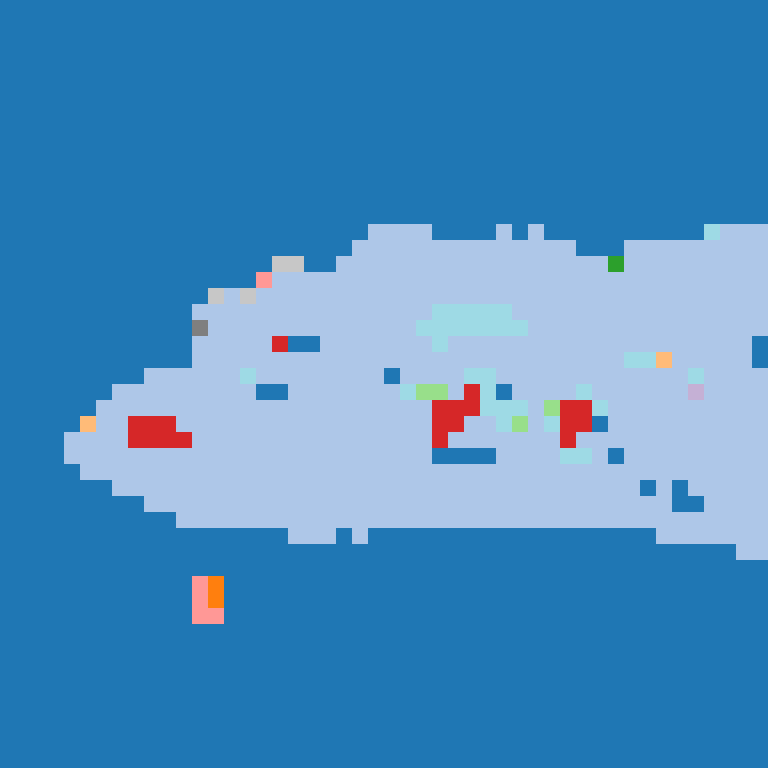} \\
            \includegraphics[width=0.32\linewidth]{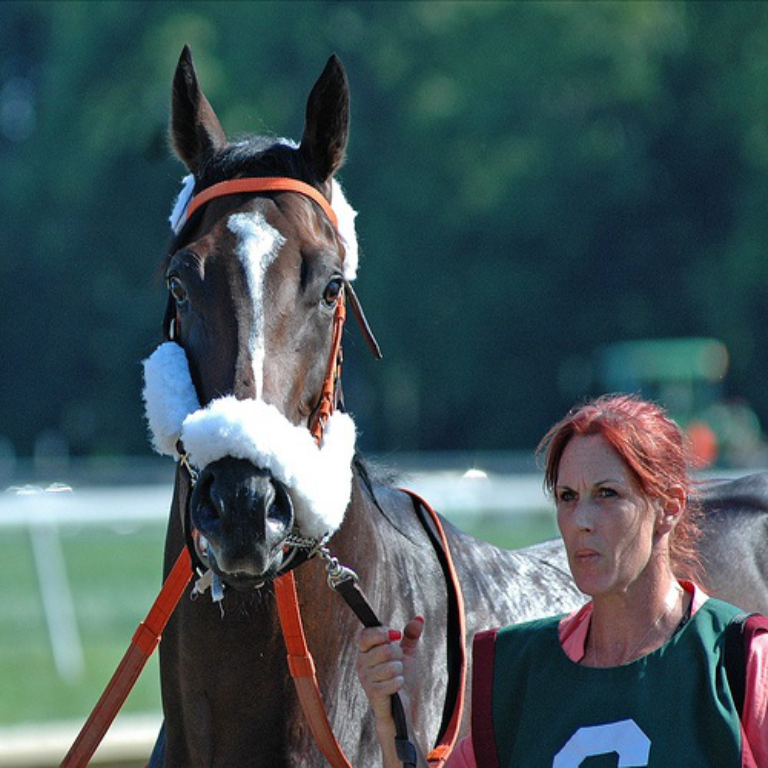} &
            \includegraphics[width=0.32\linewidth]{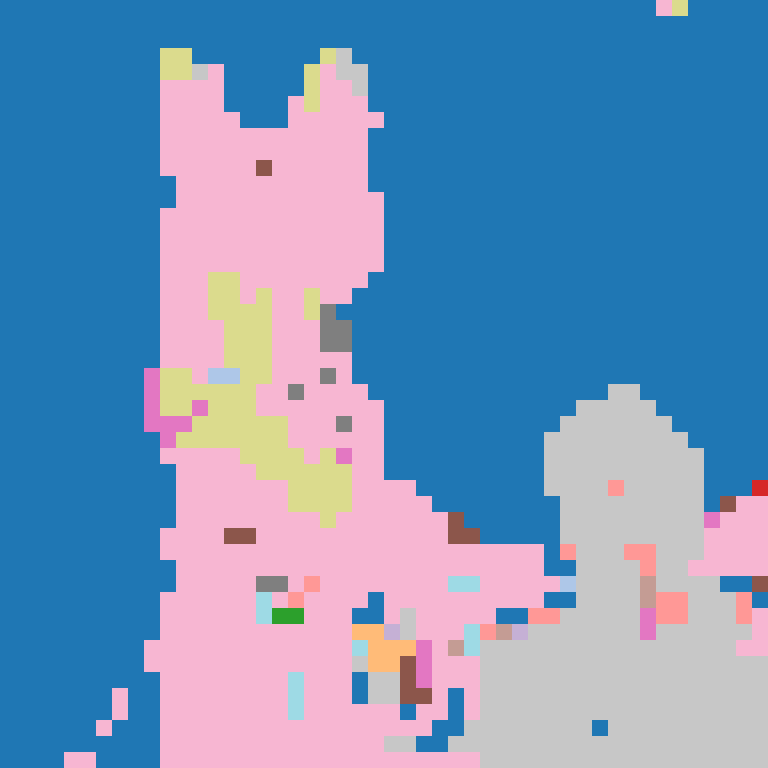} &
            \includegraphics[width=0.32\linewidth]{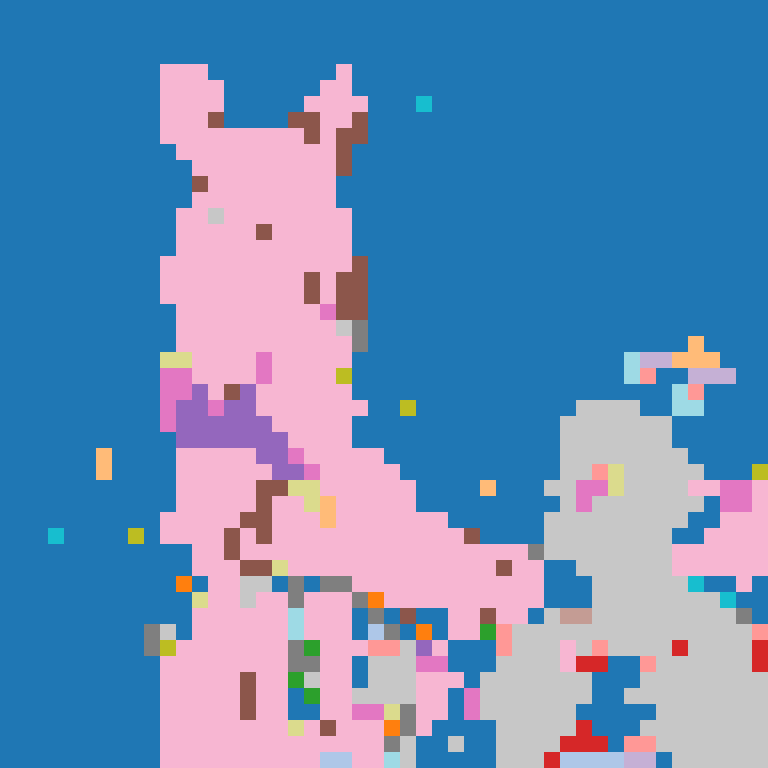}
        \end{tabular}
    \end{minipage}
    \caption{Additional qualitative results for semantic segmentation from diffusion features on Pascal VOC \cite{Everingham10}. Standard SD features use $t=100$ as the timestep, which we found to perform best quantitatively (c.f.~\suppvspreprint{Figure 8}{\cref{fig:sem_seg_quantitative}}).}
    \label{fig:add_sem_seg_qualitative}
\end{figure}
}

\begin{figure}[t]
    \centering
    \includegraphics[width=\columnwidth]{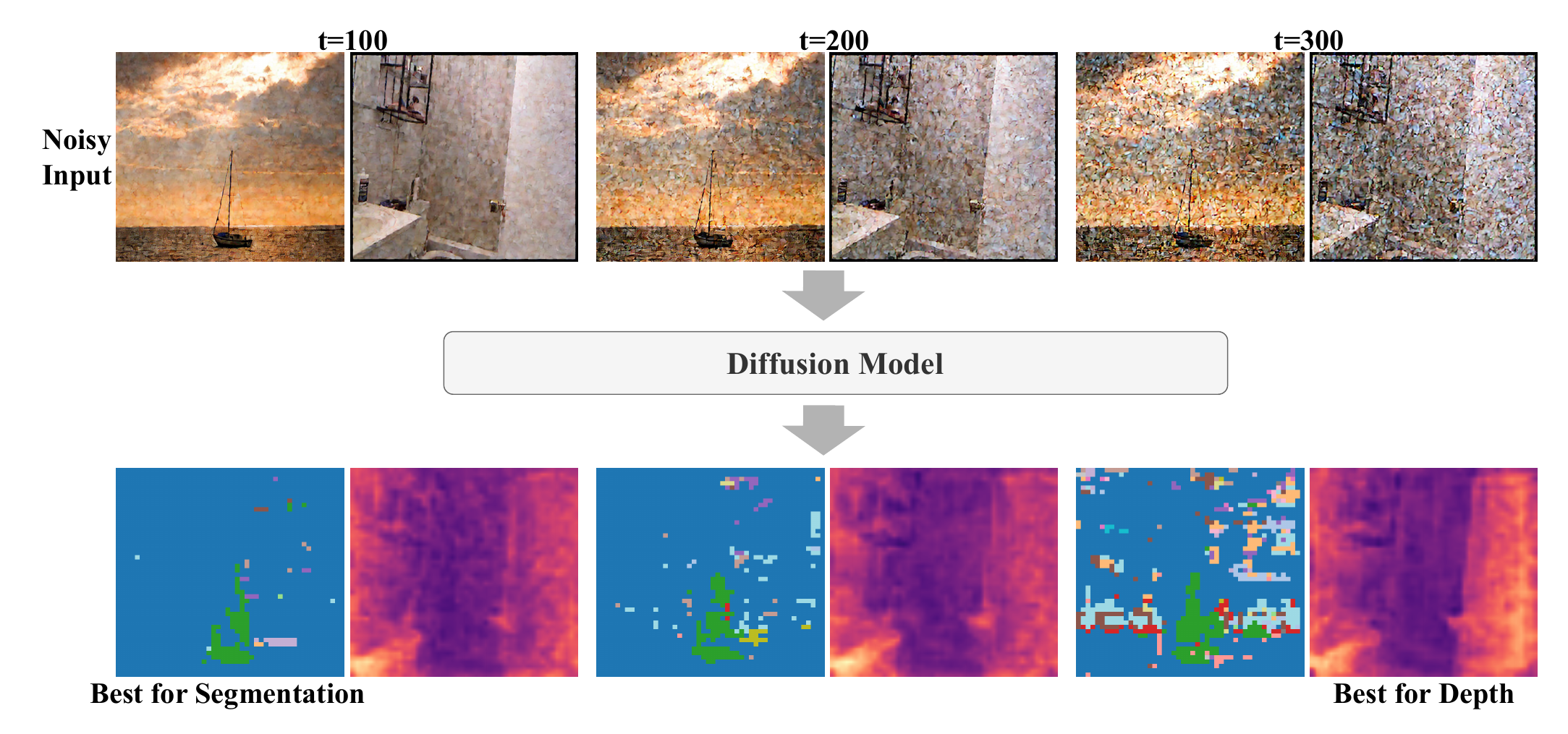}
    \caption{Depending on the downstream application, different diffusion timesteps result in optimal feature representations. For semantic segmentation, $t=100$ is optimal resulting in a much cleaner segmentation map compared to higher timesteps. However, for depth estimation, the low timestep yields inaccurate depth estimates and a higher timestep is necessary ($t=300$). CleanDIFT remedies the dependence on the timestep and yields optimal features for every downstream task without additional tuning (\cref{fig:teaser}). }
    \label{fig:teaser-fig-appendix}
\end{figure}

\begin{table}[t]
    \centering
    \vspace{2mm}
    \adjustbox{max width=\columnwidth}{
    \begin{tabular}{ccc}
        \toprule
        \small SPair71k (Test) & \small COYO (Generic, Ours) & \small ImageNet (mismatched) \\
        \midrule
         61.42 & 61.43 & 60.78 \\
        \bottomrule
    \end{tabular}
    }
    \vspace{-1.5mm}
    \caption{SPair \pck{} when training on different datasets. A generic dataset performs best and training on the test dataset does not yield any additional gains.}
    \label{tab:dataset_ablation}
\end{table}

\section{Additional Projection Head Ablations}
\label{sec:proj-heads}
We investigate the influence of different components of the projection head architecture and the influence of pre-training the projection heads following the setup in ~\suppvspreprint{Sec. 4.6}{\cref{subsec:ablation}}. In our main configuration, we use a FiLM layer~\cite{perez2018film} in each FFN block to adaptively scale activations depending on the timestep $t$. We replace the FiLM layers with Adaptive RMS (AdaRMS) norm layers~\cite{zhang2019root} and observe a performance degradation of $0.1$ percentage points for $\text{PCK}_{\text{bbox}}$. We conclude that removing the scale and shift information of the model's feature by normalization is harmful and cannot be recovered by subsequent scaling and shifting. 

Our main configuration for the projection heads uses the SwiGLU~\cite{shazeer2020glu} gating mechanism as an activation function in each FFN block. We investigate the influence of removing this gating mechanism from our FFNs, effectively leaving us with Swish layers~\cite{ramachandran2017searching}. When removing the gating mechanism, $\text{PCK}_{\text{bbox}}$ slightly decreases by $0.12$ percentage points. Additionally, we experiment with fine-tuning the projection heads before training our feature extraction model. After pre-training the projection heads, we fine-tune them in two settings: fine-tuning both the feature extraction model and projection heads, and training only the feature extraction model while locking the pre-trained projection heads. When fine-tuning both the feature extraction model and projection heads, we achieve the exact same performance as our main configuration which does not use pre-training. When locking the projection heads during fine-tuning, the performance slightly decreases by $0.16$ percentage points for $\text{PCK}_{\text{bbox}}$.

\begin{figure}[t]
    \centering
    \scalebox{.6}{\includegraphics[]{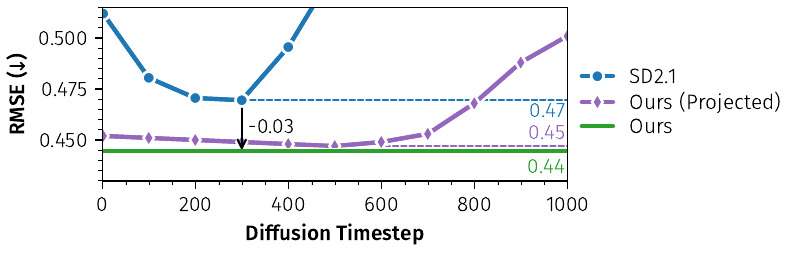}}
    \caption{Metric depth prediction for NYUv2~\cite{silberman2012nyuv2} using linear probes. We investigate our proposed projection heads' outputs by training linear probes for depth prediction on them, following the procedure described in~\suppvspreprint{Sec. 4.3}{\cref{subsec:depth}}. This figure extends the results presented in~\suppvspreprint{Tab. 2}{\cref{tab:depth_quantitative}} by showcasing the performance over timesteps. }
    \label{fig:depth-over-timesteps}
\end{figure}

\section{Other Diffusion Backbones}
We evaluate whether our method applies to other diffusion backbones by evaluating semantic correspondence performance in a DIFT~\cite{tang_emergent_2023} setting and find gains in all cases (cf.~\cref{tab:supp:more-backbones}). This shows that diffusion models suffer from noisy features independent of their size and architecture, and that CleanDIFT can successfully remedy that. Note that for SDXL (Turbo) and Flux we trained LoRAs~\cite{hu2022lora} instead of fully fine-tuning the entire model due to their large model size. We adapt the self-attention and FFN layers with LoRAs of rank 64. We swept all models for the optimal timestep and feature map and used that for the PCK calculation.

\begin{table}[t]
\centering
\adjustbox{max width=\linewidth}{
\begin{tabular}{lll}
\toprule
\multirow{2}{*}[-3pt]{Backbone} & \multicolumn{2}{c}{PCK@$\alpha$ Gain $(\uparrow)$ } \\ \cmidrule(lr){2-3} & $\alpha_\mathrm{img}$ = 0.1 & $\alpha_\mathrm{bbox}$ = 0.1   \\ \midrule
SDXL \cite{podellsdxl} & 1.7 & 1.6 \\
SDXL Turbo \cite{sauer2024adversarial} & 3.2 & 3.7 \\
PIXART-$\alpha$ \cite{chen2024pixartalpha} & 2.7 & 2.2 \\
Flux \cite{esser2024scaling} & 9.4 & 8.1 \\
\bottomrule
\end{tabular}
}
\caption{We assess the effectiveness of our features for other diffusion backbones such as a much larger UNet (SDXL), diffusion transformers (PIXART-$\alpha$, Flux), and a distilled model (SDXL Turbo). The PCK quantifies the gain obtained by using our features for semantic correspondences using the standard DIFT~\cite{tang_emergent_2023} setup when compared to the standard features. We show that using our CleanDIFT features leads to better performance for all backbones. }
\label{tab:supp:more-backbones}
\end{table}

\section{Dataset Considerations}
\label{supp:dataset}
We evaluated different datasets and find simple choices to suffice. Specifically, a random (only filtered to a minimum size of $768^2$) subset of COYO-700M with $\sim$3k images sufficed for our optimal results. COYO-700M is a dataset that is close to the original model's pretraining setting and not tailored to any of the categories relevant to our downstream evals. We show zero-shot semantic correspondence results across different fine-tuning datasets in \cref{tab:dataset_ablation}. Overfitting on the test setting, e.g., by training on SPair71k while also evaluating on that dataset does not give gains over our non-tailored version, but using a dataset that has little overlap with the distribution of the target task (e.g., ImageNet) results in reduced performance.

\begin{figure}[t]
    \centering
    \adjustbox{max width=.70\linewidth}{
        \begin{minipage}{.7\linewidth}
            \begin{tabularx}{\linewidth}{XXX}
                 \makecell[c]{Image} & \makecell[c]{DIFT \cite{tang_emergent_2023}} & \makecell[c]{Ours}
            \end{tabularx}
            \includegraphics[width=\linewidth]{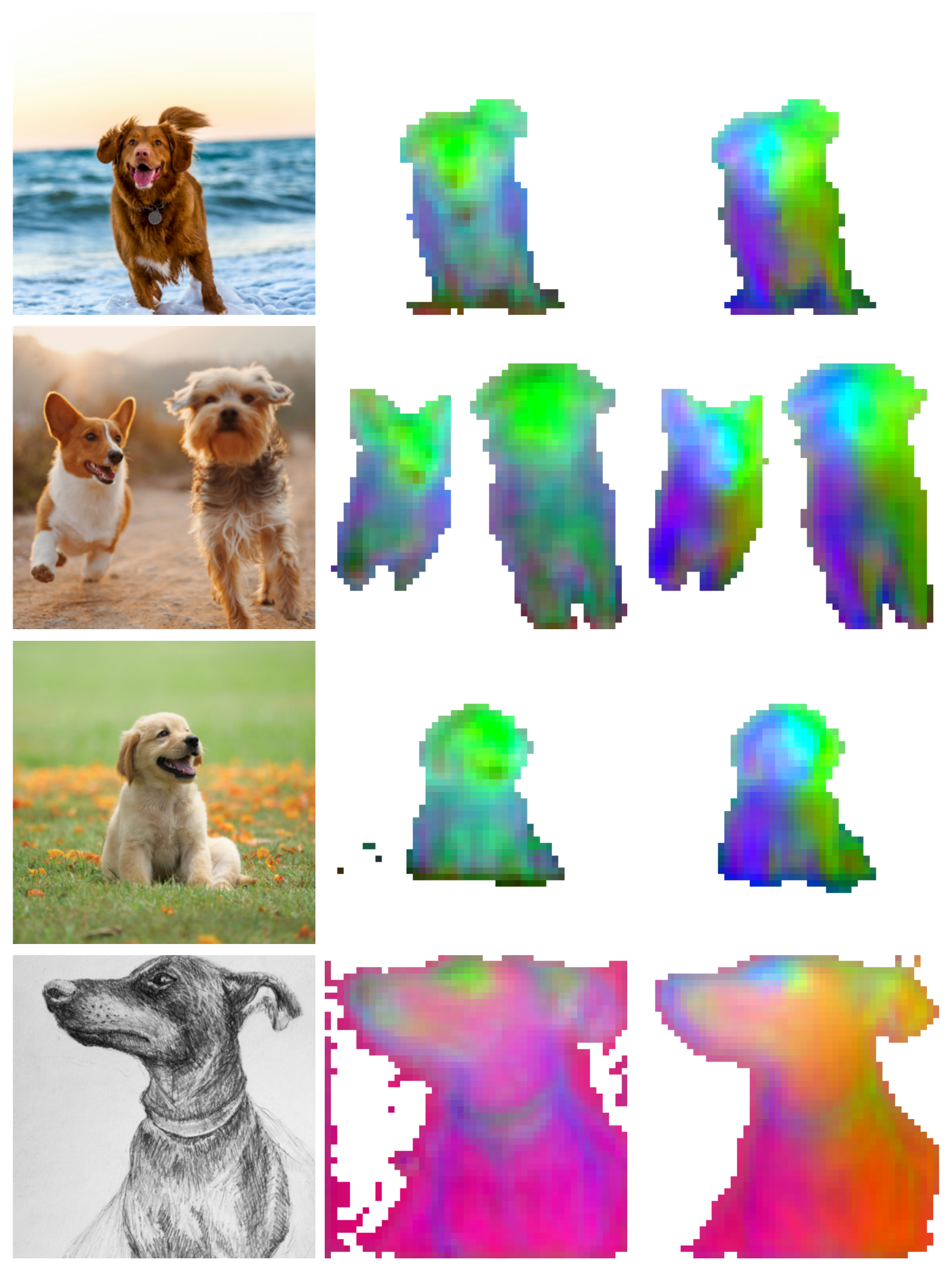}
        \end{minipage}
    }
    \caption{Principal component visualization of our features. One 4-component feature PCA is computed per column, with the first component being thresholded at 0 to obtain the foreground and the remaining being mapped to RGB color. CleanDIFT produces similarly semantically useful features as DIFT, while exhibiting less noise. }
    \label{fig:pca_qualitative}
\end{figure}

\section{Evaluation Design Choices}

During our efforts to reproduce the numbers reported by~\cite{tang_emergent_2023, zhang_tale_2023, zhang_telling_2024}, we found a variety of differences between evaluation pipelines that influence the results. We list them here to provide some clarity for future comparisons.

\paragraph{CLIP Image Embedding Conditioning} A Tale of Two Features~\cite{zhang_tale_2023} and Telling Left from Right~\cite{zhang_telling_2024} employ a conditioning mechanism on CLIP image embeddings from~\cite{xu2023open} that was fine-tuned for panoptic segmentation. Specifically, they multiply the CLIP image embedding element-wise with a learned tensor. This reweighed CLIP conditioning is then added to the embedding of an empty prompt and subsequently passed to the U-Net as the prompt embedding. Additionally, another learned tensor is used to element-wise scale the CLIP image embedding and then add it to the timestep embedding.

\paragraph{Sliding Window} Both A Tale of Two Features~\cite{zhang_tale_2023} and Telling Left from Right~\cite{zhang_telling_2024} use a sliding window approach to account for input resolutions higher than the native model input resolution. Specifically, they perform forward passes for overlapping patches, each patch having the model input resolution. Additionally, the methods use different resizing strategies to handle non-square images. DIFT~\cite{tang_emergent_2023} simply resizes the non-square input images to the square input resolution of the evaluation pipeline. A Tale of Two Features~\cite{zhang_tale_2023} and Telling Left from Right~\cite{zhang_telling_2024} resize the input image such that the longer side matches the evaluation resolution and pad the remaining part of the square image with zeros.

\end{document}